\documentclass[10pt,twocolumn,letterpaper]{article}

\usepackage[pagenumbers]{cvpr} 

\usepackage[utf8]{inputenc} 
\usepackage[T1]{fontenc}    
\usepackage{xcolor}
\usepackage{url}            
\usepackage{booktabs}       
\usepackage{amsfonts}       
\usepackage{nicefrac}       
\usepackage{microtype}      

\usepackage{graphicx}

\usepackage[most]{tcolorbox}
\usepackage{epsfig}
\usepackage{tabularx}
\usepackage[table]{xcolor}       
\definecolor{ceruleanblue}{rgb}{0.16, 0.32, 0.75}
\usepackage{relsize}
\usepackage{multirow}
\usepackage{scrextend}           
\usepackage{array}               
\usepackage{makecell}            
\usepackage{wasysym}             
\usepackage{xspace}
\usepackage{wrapfig}
\usepackage{amssymb} 
\usepackage{pifont} 
\usepackage{bm}
\usepackage{colortbl}
\usepackage[flushleft]{threeparttable}
\usepackage{subcaption}
\usepackage{pifont}
\usepackage{colortbl}
\usepackage{bm}
\usepackage{graphicx}
\usepackage{setspace}
\usepackage{cuted}
\usepackage{soul}
\usepackage{dblfloatfix}
\usepackage{esvect}
\usepackage{caption}

\definecolor{cvprblue}{rgb}{0.21,0.49,0.74}
\usepackage[pagebackref,breaklinks,colorlinks,allcolors=cvprblue]{hyperref}
\usepackage[capitalize]{cleveref}
\crefname{section}{Sec.}{Secs.}
\Crefname{section}{Section}{Sections}
\Crefname{table}{Table}{Tables}
\crefname{table}{Tab.}{Tabs.}
\usepackage[toc,page]{appendix}
\usepackage{minitoc}

\def\paperID{} 
\def\confName{CVPR}
\def\confYear{2026}

\definecolor{car}{RGB}{230,25,75}
\definecolor{cyclist}{RGB}{255, 130, 48}
\definecolor{pedestrian}{RGB}{138, 43, 226}
\definecolor{colorgt}{RGB}{170,255,195}
\definecolor{white}{rgb}{1.0, 1.0, 1.0}
\definecolor{monocop}{RGB}{71, 159, 179}
\definecolor{baseline}{RGB}{253, 190, 110}
\definecolor{lightgray}{gray}{0.95} 
\definecolor{mygreen}{RGB}{180,215,195}

\definecolor{mylightblue}{RGB}{170,200,235}

\definecolor{mypeach}{RGB}{238,200,164}

\newcommand{\monoThreeD}{Mono3D\xspace}

\newcommand{\twoD}{$2$D\xspace}
\newcommand{\threeD}{$3$D\xspace}

\newcommand{\iou}{IoU\xspace}

\newcommand{\iouThreeD}{IoU$_{3\text{D}}$\xspace}

\newcommand{\bev}{BEV\xspace}

\newcommand{\mathDash}{$-$}

\newcommand{\kitti}{KITTI\xspace}
\newcommand{\nuscenes}{nuScenes\xspace}
\newcommand{\waymo}{Waymo\xspace}

\newcommand{\valOne}{Val\xspace}

\newcommand{\val}{Val\xspace}

\newcommand{\test}{Test\xspace}

\newcommand{\ap}{AP}

\newcommand{\apThreeD}{\ap$_{3\text{D}}$\xspace}
\newcommand{\aphThreeD}{APH$_{3\text{D}}$\xspace}

\newcommand{\apBev}{\ap$_{\text{BEV}}$\xspace}

\newcommand{\bracketPercentage}{[\%]}

\newcommand{\sota}{SoTA\xspace}

\newcommand{\deviant}{DEVIANT\xspace}

\newcommand{\gupNet}{GUP Net\xspace}

\newcommand{\monodetr}{MonoDETR\xspace}
\newcommand{\monodgp}{MonoDGP\xspace}

\newcommand{\occupancymThreeD}{OccupancyM3D\xspace}
\newcommand{\opaThreeD}{OPA-3D\xspace}

\newcommand{\monocon}{MonoCon\xspace}
\newcommand{\fdThreeD}{FD3D\xspace}
\newcommand{\monoflex}{MonoFlex\xspace}

\newcommand{\monouni}{MonoUNI\xspace}
\newcommand{\monocd}{MonoCD\xspace}
\newcommand{\monomae}{MonoMAE\xspace}
\newcommand{\monocop}{MonoCoP\xspace}

\newcommand{\monotakd}{MonoTAKD\xspace}

\providecommand\rightarrowRHD{\relbar\joinrel\mathrel\RHD}

\newcommand{\uparrowRHD}  {\rotatebox[origin=c]{90}{$\rightarrowRHD$}}

\newcommand{\uparrowRHDSmall}  {\raisebox{0.05\normalbaselineskip}{\scalebox{0.7}{\uparrowRHD}}}   

\usepackage{pifont}
\newcommand{\cmark}{\ding{51}}

\newcommand{\noIndentHeading}[1]{\noindent\textbf{#1}}

\definecolor{XLcolor}{rgb}{0.858, 0.188, 0.478}

\newcommand{\apThreeDSeventy}{\ap$_{\!3\text{D}\!}$ 70\xspace}
\newcommand{\apThreeDFifty}{\ap$_{\!3\text{D}\!}$ 50\xspace}

\newcommand{\methodName}{MonoIA\xspace}

\newcommand{\methodNameFull}{Towards Intrinsic-Aware Monocular 3D Object Detection}

\title{\methodNameFull}

\author{%
  Zhihao Zhang\textsuperscript{1}\quad Abhinav Kumar\textsuperscript{1}  \quad Xiaoming Liu\textsuperscript{1,2} \\
  \textsuperscript{1}Michigan State University \quad \textsuperscript{2}University of North Carolina at Chapel Hill\\
  zhan2365@msu.edu \quad abhinav3663@gmail.com \quad liuxm@cs.unc.edu 
}

\begin{document}
\maketitle

\begin{abstract}
Monocular \threeD object detection (\monoThreeD) aims to infer object locations and dimensions in \threeD space from a single RGB image.
Despite recent progress, existing methods remain highly sensitive to camera intrinsics and struggle to generalize across diverse settings, since intrinsic governs how \threeD scenes are projected onto the image plane.
We propose \methodName, a unified intrinsic-aware framework that models and adapts to intrinsic variation through a language-grounded representation.
The key insight is that intrinsic variation is not a numeric difference but a perceptual transformation that alters apparent scale, perspective, and spatial geometry.
To capture this effect, \methodName employs large language models and vision–language models to generate intrinsic embeddings that encode the visual and geometric implications of camera parameters.
These embeddings are hierarchically integrated into the detection network via an Intrinsic Adaptation Module, allowing the model to modulate its feature representations according to camera-specific configurations and maintain consistent \threeD detection across intrinsics.
This shifts intrinsic modeling from numeric conditioning to semantic representation, enabling robust and unified perception across cameras.
Extensive experiments show that \methodName achieves new state-of-the-art results on standard benchmarks including \kitti, \waymo, and \nuscenes (\eg, $+1.18\%$ on the \kitti leaderboard), and further improves performance under multi-dataset training (\eg, $+4.46\%$ on \kitti \val). Code and models are publicly available at  \href{https://github.com/alanzhangcs/MonoIA}{https://github.com/alanzhangcs/MonoIA}. 
\end{abstract}

\section{Introduction}
\label{sec:intro}

\begin{figure}[t]
    \centering
    \includegraphics[width=0.95\linewidth]{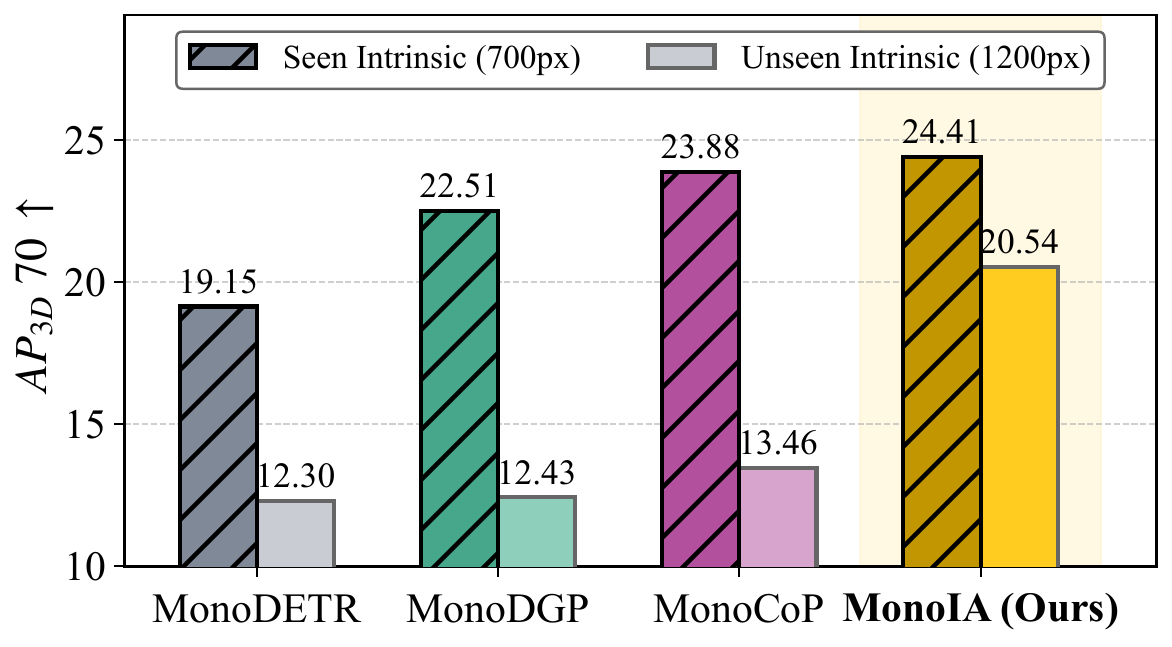}
    \vspace{-2mm}
    \caption{\textbf{\methodName enables intrinsic awareness in \monoThreeD on \kitti \val.} Existing \monoThreeD detectors~\cite{zhang2023monodetr, pu2024monodgp, zhang2025unleashing} lack intrinsic awareness and thus generalize poorly to images with unseen intrinsics. In contrast, our intrinsic-aware \methodName  achieves superior performance under seen intrinsics and demonstrates strong generalization to the unseen one. 
    }
    \label{fig:teaser}
    \vspace{-6mm}
\end{figure}

\begin{figure*}[t]
    \centering
    \includegraphics[width=0.99\linewidth]{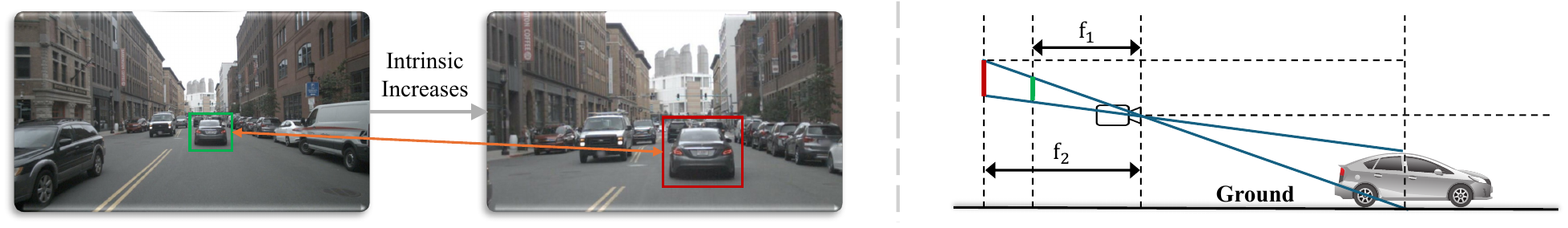}
    \vspace{-2mm}
    \caption{
        \textbf{Impact of intrinsic variation on image appearance.}
       \textbf{Left:} The two images show the same object in the same \threeD position but captured with different intrinsics. As the focal length increases, the object appears larger and the FoV is smaller. \textbf{Right:} Schematic illustration of how intrinsic variations affect object appearance.
    }
            
    \label{fig:teaser-2}
    \vspace{-3mm}
\end{figure*}

Monocular \threeD object detection (\monoThreeD) aims to estimate the \threeD locations and dimensions of objects from a single RGB image, offering a cost-effective alternative to LiDAR-based approaches~\cite{yin2021center, shi2019pointrcnn, peng2024learning}. 
Due to its low hardware requirements, \monoThreeD has attracted increasing attention in autonomous driving~\cite{simonelli2020disentangling} and robotics~\cite{ma20233d}.

In real-world scenarios, cameras exhibit diverse intrinsic parameters, making robustness to intrinsic variation essential for practical deployment. However, existing state-of-the-art (\sota) detectors~\cite{brazil2023omni3d, kumar2022deviant, yan2024monocd, pu2024monodgp, zhang2025unleashing} typically assume fixed intrinsics during both training and inference, which limits their generalization. For instance, \cref{fig:teaser} shows that \monodgp
~\cite{zhang2023monodetr}, \monodgp~\cite{pu2024monodgp} and \monocop~\cite{zhang2025unleashing} perform well when evaluated under the \emph{seen} intrinsic, but exhibit significant degradation under unseen intrinsics.
As a result, models trained under one intrinsic often fail to generalize to unseen configurations, and adapting to new cameras usually requires full retraining. 

We first expose detectors to a broader range of focal lengths by varying the field of view (FoV) of input images. However, empirical results show that such diversity alone yields only marginal gains. 
This finding reveals that the key challenge lies not in data diversity but in how detectors represent intrinsic cues. As illustrated in \cref{fig:teaser-2}, changes in focal length reshape how a \threeD scene is projected onto the image plane: the same object appears larger and the background more compressed under a longer focal length, even though its \threeD position remains unchanged. Such variations alter apparent scale, perspective, and spatial geometry, which are fundamental to reliable \monoThreeD. Yet, conventional detectors~\cite{brazil2023omni3d, pu2024monodgp, zhang2025unleashing} treat intrinsics as raw numeric inputs, forcing the network to infer their perceptual effects from limited supervision. Consequently, models tend to either ignore intrinsic cues or overfit to a few discrete training values, resulting in poor generalization to unseen configurations.

To address this gap, we propose \methodName, a unified intrinsic-aware framework that explicitly models and integrates intrinsic information throughout the detection process. \methodName introduces two key components. The \emph{Intrinsic Encoder} transforms numeric intrinsics into \textit{language-grounded} representations. For each intrinsic configuration, a large language model (LLM)~\cite{openai2024chatgpt} generates textual descriptions that capture its perceptual and geometric effects, such as changes in field of view, perspective distortion, and depth compression. These descriptions are then encoded using a CLIP Text Encoder~\cite{sun2023eva} to form semantically structured intrinsic embeddings. Unlike raw numbers, these embeddings capture how intrinsic variations manifest visually, yielding a perceptually continuous and geometrically organized representation space. This language-grounded encoding provides a strong inductive bias for intrinsic-aware feature learning and supports robust generalization to unseen focal lengths.

While the Intrinsic Encoder captures the perceptual meaning of intrinsics, the resulting embeddings remain external to the detection process. To fully leverage this knowledge, we introduce an \emph{Intrinsic Adaptation Module} that consists of a lightweight Connector and a hierarchical fusion mechanism. The fixed intrinsic embeddings are preserved to maintain their semantic consistency, while the Connector maps them into a learnable latent space for interaction with visual features. Through hierarchical fusion, these adapted intrinsic features are integrated at multiple network stages, allowing intrinsic cues to guide both low-level representation learning and high-level object reasoning. This design ensures that the semantic understanding of camera intrinsics becomes an integral part of the detection pipeline.

Overall, \methodName shifts intrinsic modeling from numeric conditioning to semantic representation. This design brings three key advantages. It improves zero shot generalization to unseen focal lengths, enables natural compatibility with multi dataset training, and delivers stronger performance on standard \threeD benchmarks, providing a unified intrinsic aware solution for robust \monoThreeD.

In summary, our main contributions are as follows:

\noindent$\bullet$ We reveal  existing \monoThreeD  methods are highly sensitive to intrinsic variations and generalize poorly to unseen intrinsics. 

\noindent$\bullet$ We identify that intrinsic variation is not a simple numeric difference but a \textit{perceptual transformation} that alters apparent scale, perspective, and spatial geometry, redefining how \threeD scenes are visually perceived.

\noindent$\bullet$ We introduce \textbf{\methodName}, a unified intrinsic-aware framework that first transforms numeric intrinsics into \textit{language-grounded representations} capturing their perceptual and geometric effects, and then integrates them \textit{hierarchically} into the detector for intrinsic-aware feature learning.

\noindent$\bullet$ Extensive experiments across multiple benchmarks demonstrate that \methodName achieves (1) superior zero-shot generalization to unseen focal lengths, (2) natural compatibility with multi-dataset training, and (3) significant accuracy gains under standard \threeD settings, validating its effectiveness and broad generalization capability.

\section{Related Work}
\label{sec:related}

\noIndentHeading{Mono3D.}
Monocular \threeD object detection (\monoThreeD) relies solely on a single RGB image as input, posing significant challenges due to the inherent ambiguity in recovering depth from \twoD projections. Early methods addressed this task using hand-crafted features~\cite{payet2011contours}, but recent advances predominantly leverage deep neural networks~\cite{brazil2020kinematic,brazil2019pedestrian}. A broad spectrum of techniques has been explored to enhance performance, including architectural improvements~\cite{huang2022monodtr,xu2023mononerd}, equivariant representations~\cite{kumar2022deviant, chen2023viewpoint}, loss function design~\cite{brazil2019m3d, chen2020monopair}, uncertainty modeling~\cite{lu2021geometry, kumar2020luvli}, and explicit depth estimation~\cite{zhang2021objects, min2023neurocs, yan2024monocd, wu2024fd3d, pu2024monodgp, kumar2025charm3r}.
Several works incorporate additional signals during training, such as non-maximum suppression (NMS)~\cite{kumar2021groomed, liu2023monocular, zhu2020edge}, corrected extrinsics \cite{zhou2021monoef}, CAD models~\cite{chabot2017deep, liu2021autoshape, lee2023baam}, or even LiDAR supervision~\cite{reading2021categorical, huang2024training, long2023radiant, long2025riccardo}. Others propose innovations like pseudo-LiDAR representations~\cite{wang2019pseudo, ma2019accurate, huang2024training}, diffusion-based generation~\cite{ranasinghe2024monodiff, lin2025drivegen}, or BEV (bird’s-eye view) encoding~\cite{jiang2024fsd, zhang2022beverse, li2024bevnext}. 
Transformer-based approaches~\cite{zhang2023tile} have also gained traction~\cite{carion2020detr}, with modifications including positional encoding~\cite{shu2023dppe, tang2024simpb, hou2024open}, learned queries~\cite{li2023fast, zhang2023dabev, ji2024enhancing, chen2024learning}, and query denoising~\cite{liu2024ray}.
Additional techniques include knowledge distillation~\cite{wang2023distillbev, kim2024labeldistill, liu2025monotakd, yang2025monoclue}, stereo input~\cite{wang2022sts, li2023bevstereo}, and advanced loss functions~\cite{kumar2024seabird, liu2024multi}. For a broader overview of the field, we refer readers to recent surveys~\cite{ma20233d, ma2022vision}. Our \methodName focuses on intrinsic-aware \monoThreeD, improving performance across unseen intrinsics.

\begin{figure*}[htb]
    \centering
    \includegraphics[width=1\linewidth]{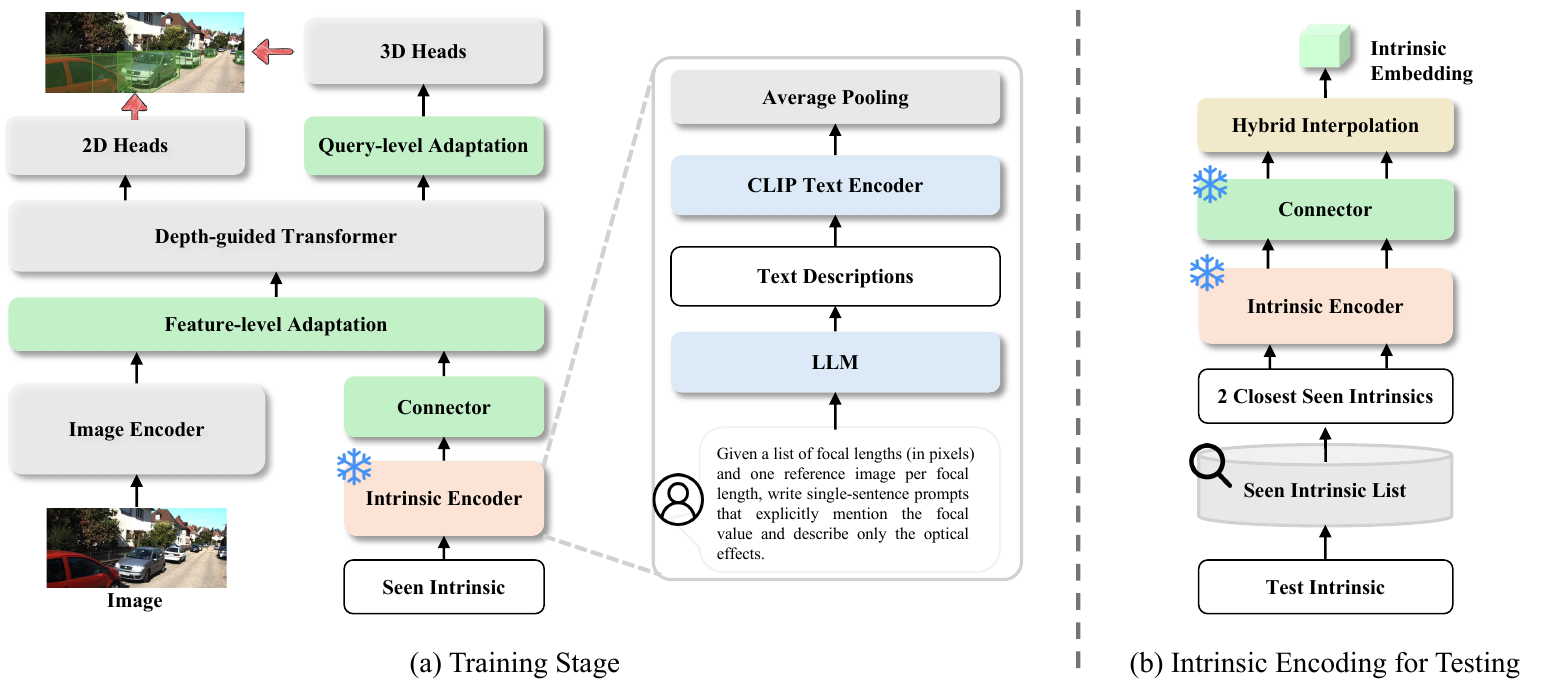}
    \vspace{-6mm}
    \caption{
    \textbf{Overview of \methodName.} 
    (a) \textbf{Training Stage:} 
    \methodName is a unified intrinsic-aware detection framework built upon two designs. 
    The \textcolor[HTML]{F5A572}{\textbf{Intrinsic Encoder}} leverages the knowledge of LLM and CLIP to convert numeric intrinsics into semantically meaningful embeddings that capture their perceptual and geometric effects, providing a strong prior for generalization across cameras. 
    The \textcolor[HTML]{8DC88F}{\textbf{Intrinsic Adaptation Module}} bridges this semantic knowledge with visual perception through a lightweight Connector and hierarchical fusion, enabling the detector to interpret visual features in an intrinsic-aware manner and maintain consistent \threeD detection under diverse camera settings.
    (b) \textbf{Testing Stage:} 
    For each test intrinsic, we retrieve its two nearest seen intrinsics together with their embeddings, and then apply a \textcolor[HTML]{E8D78A}{\textbf{Hybrid Interpolation Strategy}} that adaptively switches between nearest-neighbor selection and linear interpolation. If the intrinsic gap is $\leq 32$ px, the nearest seen embedding is reused; otherwise, the two nearest embeddings are linearly interpolated to synthesize the test intrinsic embedding.
    }
    \label{fig:monoia-framework}
    \vspace{-3mm}
\end{figure*}

\noIndentHeading{Foundation Models in \threeD Tasks.}
Foundation models~\cite{oquab2023dinov2} such as Large Language Models (LLMs)~\cite{qwen} and vision-language models like CLIP~\cite{sun2023eva, Radford2021LearningTV} have demonstrated remarkable semantic understanding and cross-modal alignment, driven by large-scale pretraining on massive text and \twoD image datasets~\cite{schuhmann2022laionb}.
However, due to the larger search space in \threeD and the limited availability of large-scale \threeD datasets~\cite{geiger2012we}, analogous foundation models for \threeD tasks are still lacking.
Recent efforts~\cite{Peng2023OpenScene} aim to bridge this gap by leveraging existing \twoD or language foundation models to enhance \threeD performance.
For instance, some~\cite{zhang2024tamm, xue2023ulip} pretrain \threeD encoders under CLIP supervision, while others~\cite{hong20233d, xu2024pointllm} utilize LLMs for reasoning in complex \threeD scenes.
Meanwhile, CLIP Text Encoder has been extensively used in vision and generation tasks~\cite{esser2024scaling, yuan2025generative, shi2023zero123++}, demonstrating its ability to encode fine-grained textual semantics, numeric descriptions, and geometric attributes.
Distinct from these prior efforts, our work is the first to leverage LLM and CLIP for encoding \emph{camera intrinsics}, transforming numeric parameters into semantically meaningful embeddings that enhance robustness and generalization in \monoThreeD.

\section{Approach}
\label{sec:method}

\noindent\textbf{Overview.}
As \cref{fig:monoia-framework} shows, \methodName comprises three components that jointly simulate, represent, and integrate camera intrinsics within the detection framework. 
The \emph{Intrinsic Simulation Module} (see~\cref{sec:data-curation}) generates images with diverse focal lengths while preserving geometric consistency, enriching the training distribution. 
The \emph{Intrinsic Encoder Module} (see~\cref{sec:intrinsic-encoder}) leverages large language and vision–language models to transform numeric intrinsics into semantic representations that capture their perceptual and geometric effects. 
Finally, the \emph{Intrinsic Adaptation Module} (see~\cref{sec:intrinsic-fusion}) injects these embeddings into the detector through a lightweight Connector and hierarchical fusion, enabling consistent \threeD understanding across cameras with varying intrinsics.

\begin{figure}[t]
    \centering
    \includegraphics[width=0.98\linewidth]{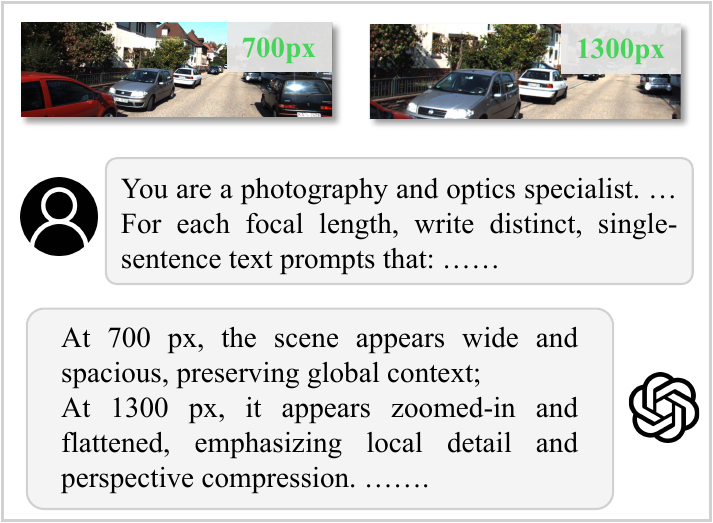}
    \caption{\textbf{LLM-Guided Description Generation.} 
    Images rendered with diverse camera intrinsics are fed into an LLM, which generates concise descriptions linking each intrinsic’s perceptual and geometric effects with its numeric focal value, forming semantic intrinsic descriptions.}
    \label{fig:text-generation}
    \vspace{-3mm}
\end{figure}

\subsection{Intrinsic Simulation Module}
\label{sec:data-curation}

Prior works handle intrinsic variation by normalizing all images to a canonical focal length~\cite{brazil2023omni3d} or by applying heuristic \twoD augmentations such as random cropping and scaling~\cite{kumar2022deviant}. 
These strategies either eliminate intrinsic diversity or distort the geometric relationship between focal length and field of view (FoV).
To address this, we design an \emph{Intrinsic Simulation Module} that performs \textit{FoV-based image approximation} to emulate diverse focal lengths while preserving geometric plausibility. 
Given an image and its intrinsic matrix $\mathbf{K}_{\text{orig}}$, we randomly sample a target focal length $f_i \in [700, 1300]$ and compute the corresponding FoV:
\begin{equation}
\theta = 2 \arctan\left( \frac{w}{2f_i} \right),
\end{equation}
where $w$ denotes the image width. 
A smaller $f_i$ yields a wider FoV (zoom-out effect), while a larger $f_i$ produces a narrower FoV (zoom-in effect). 
The simulated image is obtained by resizing the original according to the new FoV, effectively mimicking different camera perspectives without any \threeD re-rendering or depth supervision. 
Although approximate, this lightweight transformation efficiently increases intrinsic diversity while maintaining geometric consistency, allowing the detector to experience a wide spectrum of camera configurations and preparing it for intrinsic-aware learning. We provide  simulated image samples in Appendix\,\ref{appx:syn-images}.

\subsection{Intrinsic Encoder}
\label{sec:intrinsic-encoder}

While the Intrinsic Simulation Module exposes detectors to diverse focal lengths, data diversity alone is insufficient for achieving intrinsic awareness. 
Empirically, directly training detectors such as \monocop~\cite{zhang2025unleashing} on simulated images yields marginal gains, indicating that raw numeric intrinsics (\eg, focal length) provide weak inductive bias. 
These values do not convey how intrinsic changes alter perceived geometry, scale, or perspective, which are essential cues for intrinsic-aware reasoning. 
To bridge this gap, we introduce an \emph{Intrinsic Encoder} (see~\cref{fig:monoia-framework}a) that maps numeric intrinsics into \textit{language-grounded} representations, enabling the detector to interpret intrinsics through their perceptual and geometric implications.

\noindent\textbf{LLM-Guided Description Generation.}
As shown in~\cref{fig:text-generation}, for each focal length $f_i$, an LLM receives its numeric value and a simulated image from the Intrinsic Simulation Module, and generates $N$ concise, content-independent descriptions that capture the optical effects of this intrinsic setting (\eg, changes in field of view, perspective distortion, and depth compression).  
This prompting design explicitly ties quantitative intrinsics to perceptual outcomes, allowing the LLM to express how focal variation reshapes visual appearance.  
For example, a shorter focal length yields a wide and spacious view emphasizing global context, whereas a longer focal length compresses perspective and magnifies distant objects.
We provide additional details of the prompts and the generated text descriptions in the Appendix\,\ref{appx:llm}.

\noindent\textbf{Text Encoding and Embedding Formation.}
The generated intrinsic descriptions $\{p_i\}_{i=1}^{N}$ are encoded into text embeddings by CLIP Text Encoder~\cite{Radford2021LearningTV}:
\begin{equation}
\mathbf{t}_i = \text{CLIP}_{\text{Text}}(p_i), \quad \mathbf{t}_{\text{avg}} = \frac{1}{N}\sum_{i=1}^{N}\mathbf{t}_i.
\end{equation}
Averaging across descriptions yields an intrinsic embedding $\mathbf{t}_{\text{avg}}$ that captures the shared perceptual meaning of each focal length.  
CLIP encodes the LLM-generated descriptions into a semantic space where numerically close focal lengths map to similar embeddings, forming a perceptually continuous and geometry-aware representation.

\begin{figure}[t]
        \centering
        \includegraphics[width=\linewidth]{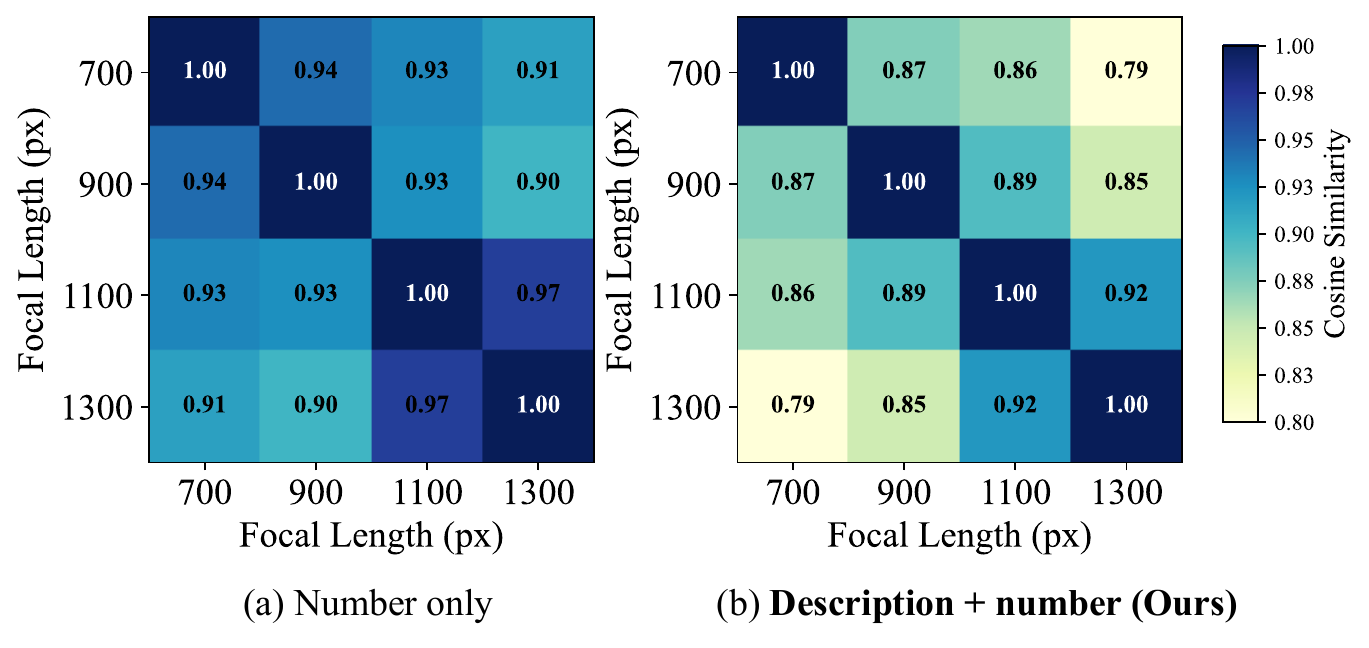}
        \vspace{-6mm}
        \caption{\textbf{Cosine similarity of intrinsic embeddings under different encoding strategies} (a) Numeric-only encoding produces uniformly high similarity, showing that CLIP text embeddings of raw focal values lack discriminative structure.
        (b) Our Intrinsic Encoder, which integrates LLM-generated perceptual descriptions with numeric grounding, yields a smooth and ordered similarity pattern, indicating a structured and geometry-aware intrinsic space.}
        \label{fig:heatmap}
        \vspace{-3mm}
\end{figure}

\noindent\textbf{Embedding Analysis.}
We visualize pairwise cosine similarities among embeddings for focal lengths between 700–1300.  
As shown in~\cref{fig:heatmap}, numeric-only encodings produce uniformly high similarity, indicating a lack of geometric structure.  
In contrast, our language-guided CLIP embeddings exhibit an ordered pattern where neighboring focal lengths are more correlated, demonstrating that the Intrinsic Encoder successfully models focal variation.

\subsection{Intrinsic Adaptation Module}
\label{sec:intrinsic-fusion}

While the Intrinsic Encoder produces semantically rich embeddings that capture the perceptual and geometric meaning of camera intrinsics, these embeddings remain external to the detection process. 
To make intrinsic awareness actionable, we introduce an \emph{Intrinsic Adaptation Module} that integrates intrinsic embeddings into the \monoThreeD through a lightweight \emph{Connector} and a hierarchical fusion mechanism. 
The Connector bridges the frozen semantic space and the learnable visual space, while the hierarchical fusion injects intrinsic cues into both feature maps and transformer queries, enabling the detector to adapt its representations according to camera geometry.

\noindent\textbf{Bridging Semantic and Visual Spaces.}
Intrinsic embeddings from the Intrinsic Encoder reside in a high-level language-aligned space. 
To preserve their semantic priors while allowing task-specific adaptation, the Connector projects these frozen embeddings into a trainable, vision-aligned space using a two-layer MLP with GELU activation~\cite{hendrycks2016gaussian}. 
This projection serves as an interface between semantic priors and visual features, ensuring that intrinsic cues can modulate the detection process without disrupting their original structure and semantic  meaning.

\noindent\textbf{Hierarchical Intrinsic Fusion.}
We then hierarchically inject the transformed intrinsic embedding $\mathbf{t}_{\text{intr}}$ into the detector at both the feature and query levels.

\noindent\emph{(a) Feature-Level Adaptation.}  
At early stages, intrinsic information conditions the multi-scale backbone features on camera geometry. 
Given feature maps $\mathbf{F}_1$, $\mathbf{F}_2$, and $\mathbf{F}_3$, each is projected to a shared dimension $d'$ via a $1{\times}1$ convolution. 
The intrinsic embedding is broadcast and added to each spatial position:
\begin{equation}
\widetilde{\mathbf{F}}_i(x, y) = \mathbf{F}_i'(x, y) + \mathbf{t}_{\text{intr}}, \quad i = 1, 2, 3.
\end{equation}
This conditioning injects camera awareness into the feature hierarchy, allowing the backbone to maintain geometric consistency across different intrinsics.

\noindent\emph{(b) Query-Level Adaptation.}  
To propagate intrinsic context into object-level prediction, we modulate object queries used for \threeD prediction as:
\begin{equation}
\widetilde{\mathbf{q}}_j = \mathbf{q}_j + \mathbf{t}_{\text{intr}}, \quad j = 1, 2, \dots, N_q.
\end{equation}
Each query $\mathbf{q}_j$ corresponds to a potential object hypothesis whose appearance and projection depend on the camera intrinsics. 
This fusion enables the decoder to interpret visual evidence under different focal configurations, producing more stable depth estimation and consistent \threeD localization across cameras.
Overall, the Intrinsic Adaptation Module links semantic understanding of camera intrinsics with \threeD understanding, effectively turning intrinsic knowledge into \monoThreeD detection.

\subsection{Loss Function and Inference}

\noindent\textbf{Training.} During training, the Intrinsic Encoder is frozen to maintain its pre-trained semantic space, while the Intrinsic Adaptation Module is trained jointly with the detector. Following DETR-based approaches~\cite{carion2020end, zhang2025unleashing}, \methodName uses the Hungarian algorithm to match predictions with ground-truth annotations. 
The overall training loss is defined as:
\begin{equation} 
\mathcal{L}_{\text{overall}} = \frac{1}{N_{gt}} \sum_{n=1}^{N_{gt}} \left( \mathcal{L}_{2D} + \mathcal{L}_{3D} + \mathcal{L}_{\text{dmap}} \right),
\end{equation}
where $N_{gt}$ is the number of ground-truth objects. 
$\mathcal{L}_{2D}$ denotes the \twoD bounding box loss, $\mathcal{L}_{3D}$ supervises \threeD attributes, and $\mathcal{L}_{\text{dmap}}$ corresponds to the object-level depth map prediction loss~\cite{zhang2023monodetr}.

\noindent\textbf{Inference.}
During testing, as illustrated in~\cref{fig:monoia-framework}b, \methodName performs intrinsic-aware prediction without any retraining. 
For each test image, we retrieve its two nearest seen intrinsics and their corresponding embeddings from the frozen Intrinsic Encoder and Connector. 
A \emph{Hybrid Interpolation Strategy} is then applied to synthesize the target intrinsic embedding:  
if the focal difference is within $32$\,px, the nearest embedding is reused; otherwise, the two nearest embeddings are linearly interpolated.  
The $32$\,px threshold corresponds to the smallest perceivable change after the backbone’s $32\times$ spatial downsampling, where finer focal variations become indistinguishable in the feature space.  
The synthesized intrinsic embedding is finally injected into the Intrinsic Adaptation Module to modulate visual features, ensuring consistent and robust \threeD detection under unseen camera intrinsics.

\section{Experiments}
\label{sec:experiment}

\begin{table*}[t]
    \centering
    \tabcolsep=0.12cm
    \resizebox{1\textwidth}{!}{
    \begin{tabular}{lccccccccccccccccc}
    \toprule[1pt]
    \midrule
    \multirow{2}{*}{Method}  &  \multicolumn{4}{c}{\textbf{Seen Focals (px)}} & \multicolumn{13}{c}{\textbf{Unseen Focals (px)}} \\ 
    
    &  \cellcolor{mygreen}700 & \cellcolor{mygreen}900 & \cellcolor{mygreen}1100 & \cellcolor{mygreen}1300 & \cellcolor{mypeach}600 & \cellcolor{mypeach}650 & \cellcolor{mylightblue}750 & \cellcolor{mylightblue}800 & \cellcolor{mylightblue}850 & \cellcolor{mylightblue}950 & \cellcolor{mylightblue}1000 & \cellcolor{mylightblue}1050 & \cellcolor{mylightblue}1150 & \cellcolor{mylightblue}1200 & \cellcolor{mylightblue}1250 & \cellcolor{mypeach}1350 & \cellcolor{mypeach}1400 \\
    \midrule
    \monodetr~\cite{zhang2023monodetr} & 19.15	& 18.90	& 16.76	& 14.22 & 14.09 & 16.67 & 18.55	& 17.89	& 16.21 & 16.54 & 15.12	& 15.06	& 13.66	& 12.30 & 11.88 & 10.08 & 7.51\\
    \monodgp~\cite{pu2024monodgp} & 22.51	& 21.04	& 19.96	& 16.74 & 17.42 & 19.28  &19.78	& 19.07 & 18.51 & 17.33 & 16.03 & 15.63 & 13.18 & 12.43 & 12.47 & 10.27 & 7.56\\
    \monocop~\cite{zhang2025unleashing}  & {23.88}	& {23.30}	& {22.59}	& {18.50} & 18.18 & 21.70 &  {22.49}	& {21.44}	& {20.20}	& {18.61} & {17.69} & {16.43} & {14.57} & {13.46} & {13.11} & 12.73 & 11.11\\ 
    \textbf{\methodName (Ours)}   & \textbf{24.41}	& \textbf{24.36}	& \textbf{23.69}	& \textbf{21.20} & \textbf{22.43} & \textbf{23.41}& \textbf{24.13}	& \textbf{22.93} & \textbf{23.64} & \textbf{22.48} &	\textbf{22.65} & \textbf{22.52} & \textbf{19.07} & \textbf{20.54} & \textbf{20.80} & \textbf{19.25} & \textbf{16.99}\\
    \midrule
    \bottomrule[1pt]
    \end{tabular}}
    \vspace{-2mm}
\caption{\textbf{Results on seen and unseen focal lengths.}
\sethlcolor{mygreen}\hl{Seen focals} include 700, 900, 1100, and 1300~px. 
Unseen focals include \sethlcolor{mylightblue}\hl{interpolated focals} that lie within the training interval and 
\sethlcolor{mypeach}\hl{extrapolated focals} that extend beyond the training range. 
\methodName achieves the highest \apThreeD across all focal lengths and maintains strong robustness even under extrapolated intrinsics.}

    \label{exp:seen_and_unseen_focal}
    \vspace{-2mm}
\end{table*}

\subsection{Experimental Settings}
We evaluate \methodName through three complementary settings designed to assess its generalization, scalability, and benchmark performance.
First, we examine \emph{zero-shot generalization} on \kitti~\cite{geiger2012we} using synthetic intrinsic variations generated by our Intrinsic Simulation Module.
Second, we investigate \emph{multi-dataset training} on the combination of KITTI, \nuscenes~\cite{caesar2020nuscenes}, and \waymo~\cite{sun2020scalability}, which measures the ability of \methodName to unify data with heterogeneous intrinsic configurations.
Finally, we report results on standard benchmarks, including \kitti, \nuscenes, and \waymo, to verify that intrinsic awareness not only improves cross-intrinsic robustness but also enhances accuracy under conventional evaluation protocols.

\noindent \textbf{Evaluation Metrics.}
 We report \apThreeD{} and \apBev{} using IoU thresholds of $0.7$ (Car) and $0.5$ (Pedestrian, Cyclist)\cite{simonelli2019disentangling} for \kitti.
On \waymo, we use the \aphThreeD{} metric\cite{reading2021categorical} and report results for three distance ranges: $[0,30)$, $[30,50)$, and $[50,\infty)$ meters.
On \nuscenes, we follow~\cite{zhang2025unleashing} and adopt KITTI style metrics for simplicity and consistency.

\noindent \textbf{Implementation Details.}
\methodName is built on \monocop~\cite{zhang2025unleashing}. We employ ChatGPT-4o~\cite{openai2024chatgpt} to generate text  prompts per intrinsic and adopt CLIP ViT-H/14~\cite{Radford2021LearningTV} Text Encoder. 
We design two training settings. For \textit{single-dataset} training, we train for $250$ epochs on one NVIDIA A6000 GPU with a batch size of $16$ and a learning rate of $2\times10^{-4}$ using AdamW (weight decay $10^{-4}$). For \textit{multi-dataset} training, we train for $120$ epochs on four NVIDIA A6000 GPUs with the same batch size and learning rate. 
Additional implementation details are provided in the Appendix\,\ref{appx:implementation}.

\begin{figure}[t]
        \centering
        \includegraphics[width=\linewidth]{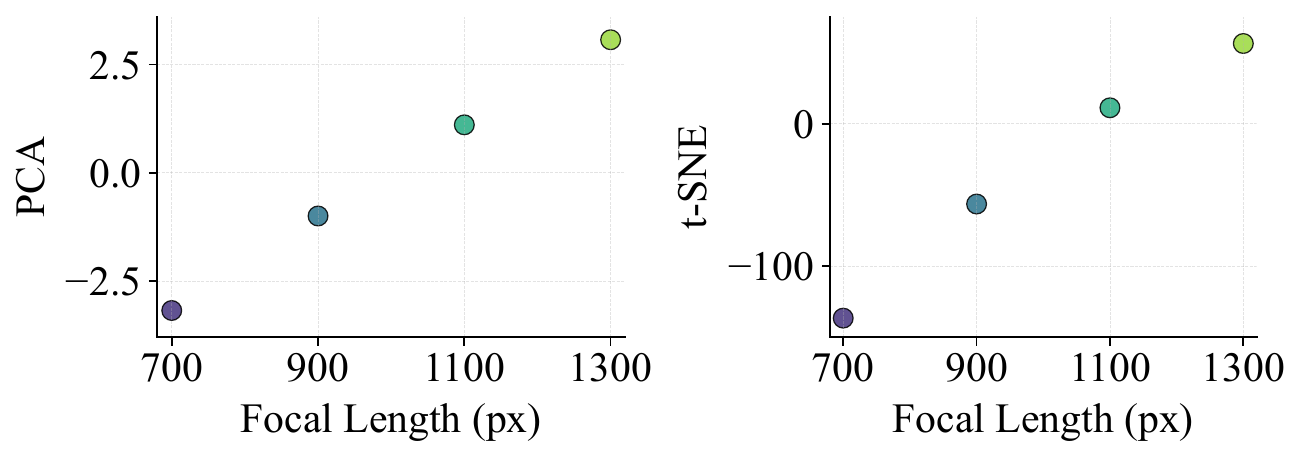}
        \vspace{-3mm}
        \caption{\textbf{Visualization of learned intrinsic embeddings.} PCA (Left) and t-SNE (Right). Both views exhibit a smooth, ordered distribution along focal length, indicating that the intrinsic embedding space learned by \methodName is geometrically consistent and well structured, facilitating interpolation for unseen intrinsics.}
        \label{fig:dimenstion}
        \vspace{-3mm}
\end{figure}

\subsection{Generalization on Synthetic Intrinsics} 

\noindent \textbf{Analysis of Learned Intrinsic Embeddings.}
\methodName learns four intrinsic embeddings corresponding to focal lengths $(700, 900, 1100, 1300)$. 
To examine whether these embeddings capture intrinsic variation, we visualize them using Principal Component Analysis (PCA) and t-Distributed Stochastic Neighbor Embedding (t-SNE). 
As shown in~\cref{fig:dimenstion}, PCA reveals a clear monotonic trajectory aligned with focal length, while t-SNE forms well-separated clusters, confirming that intrinsic embeddings preserve geometric relationships across cameras. 
Such structured embedding continuity supports our \textit{Hybrid Interpolation Strategy}, enabling unseen intrinsics to be synthesized via interpolation.

\noindent \textbf{Results on Seen Intrinsics.}
Existing \monoThreeD\ detectors are typically trained under a fixed intrinsic and generalize poorly across cameras. 
For a fair comparison, we train each baseline (\monodetr, \monodgp, and \monocop) individually under each focal length, while \methodName is trained jointly on all four intrinsics, with random sampling to ensure equal total training exposure. 
Despite this more challenging multi-focal setting, \methodName achieves the best performance across  all seen focal lengths (see~\cref{exp:seen_and_unseen_focal}), showing that intrinsic-aware modeling enhances both efficiency and accuracy.

\begin{table}[t]
\centering
\resizebox{0.88\linewidth}{!}{
\begin{tabular}{lcccc}
\toprule[1pt]
\midrule
Method 
& GT 
& $\pm5$~px 
& $\pm10$~px 
& $\pm15$~px \\
\midrule
\monodetr~\cite{zhang2023monodetr} 
& 18.55 & 16.89 & 14.95 & 11.21 \\

\monodgp~\cite{pu2024monodgp} 
& 19.78 & 17.76 & 15.38 & 12.66 \\

\monocop~\cite{zhang2025unleashing} 
& 22.49 & 20.53 & 19.22 & 15.42 \\

\textbf{\methodName (Ours)} 
& 24.13 & 23.88 & 22.34 & 18.98 \\
\midrule
\bottomrule[1pt]
\end{tabular}}
\caption{\textbf{Results under intrinsic mismatch with different perturbation levels.}
We evaluate performance under the ground truth and perturbed focal lengths ($\pm5$, $\pm10$, $\pm15$ px).}
\label{tab:intrinsic_mismatch}
\vspace{-3mm}
\end{table}

\noindent\textbf{Results on Unseen Intrinsics.}
To comprehensively evaluate intrinsic generalization, we analyze four aspects:
(1) interpolation within the training interval, 
(2) extrapolation beyond the training interval,
(3) sensitivity to intrinsic mismatch.

\noindent\textit{(1) Interpolation.}
The model is trained on focal lengths $(700, 900, 1100, 1300)$ and tested on intermediate values.
As shown in~\cref{exp:seen_and_unseen_focal}, \methodName consistently achieves the highest \apThreeD and remains stable across all interpolated focals.

\noindent\textit{(2) Extrapolation}.
We further evaluate focal lengths outside the training range, including values smaller than $700$ or larger than $1300$.
Although extrapolation is naturally more challenging than interpolation, \methodName still delivers clearly superior performance compared with all baselines, demonstrating strong robustness to unseen intrinsic configurations.

\noindent\textit{(3) Sensitivity to Intrinsic Mismatch.}
So far, all experiments assume access to the ground truth intrinsic parameters for each test sample during inference.
However, in real world applications this assumption may not be true due to calibration error~\cite{zhu2023tame}. 
To examine the behavior of MonoIA when the provided intrinsics deviate from the ground truth, we perturb the input focal length by $\pm5$, $\pm10$, and $\pm15$ px during inference.
For each perturbation magnitude, we report the average accuracy obtained under the perturbed intrinsics. As shown in \cref{tab:intrinsic_mismatch}, \methodName consistently exhibits the smallest performance drop across all perturbation levels, while existing baselines deteriorate rapidly as the mismatch increases. These results indicate that our intrinsic-aware design offers improved robustness to  miscalibrated intrinsics.

\vspace{-1mm}
\begin{table*}[t]
    \centering
    \tabcolsep=0.15cm
    \resizebox{0.98 \textwidth}{!}{  %
        \begin{tabular}{lcccccccccccccc}
            \toprule[1pt]
            \midrule
            \multirow{2}{*}{Method} & \multirow{2}{*}{\makecell{Extra \\ Data}}  
            & \multicolumn{3}{c}{\test, \apThreeD (\uparrowRHDSmall)} 
            & \multicolumn{3}{c}{\test, \apBev (\uparrowRHDSmall)} 
            & \multicolumn{3}{c}{\val, \apThreeD (\uparrowRHDSmall)} 
            & \multicolumn{3}{c}{\val, \apBev (\uparrowRHDSmall)} \\
            
            & & Easy & Mod. & Hard & Easy & Mod. & Hard 
              & Easy & Mod. & Hard & Easy & Mod. & Hard \\ 
            \midrule

            \occupancymThreeD~\cite{peng2024learning} & LiDAR  
            & 25.55 & 17.02 & 14.79 & 35.38 & 24.18 & 21.37 
            & 26.87 & 19.96 & 17.15 & 35.72 & 26.60 & 23.68 \\

            \opaThreeD~\cite{su2023opa} & Depth  
            & 24.68 & 17.17 & 14.14 & 32.50 & 23.14 & 20.30 
            & 24.97 & 19.40 & 16.59 & 33.80 & 25.51 & 22.13 \\

            \monotakd~\cite{liu2025monotakd} & LiDAR  
            & 27.91 & 19.43 & 16.51 & 38.75 & 27.76 & 24.14 
            & 34.36 & 22.61 & 19.88 & 42.86 & 29.41 & 26.47 \\
                        
            \midrule

            \monouni~\cite{jia2023monouni} & None  
            & 24.75 & 16.73 & 13.49 & \mathDash & \mathDash & \mathDash  
            & 24.51 & 17.18 & 14.01 & \mathDash  & \mathDash  & \mathDash \\

            \monodetr~\cite{zhang2023monodetr} & None 
            & 25.00 & 16.47 & 13.58 & 33.60 & 22.11 & 18.60 
            & 28.84 & 20.61 & 16.38 & 37.86 & 26.95 & 22.80 \\

            \monocd~\cite{yan2024monocd} & None 
            & 25.53 & 16.59 & 14.53 & 33.41 & 22.81 & 19.57 
            & 26.45 & 19.37 & 16.38 & 34.60 & 24.96 & 21.51 \\

            \monomae~\cite{jiang2024monomae} & None 
            & 25.60 & 18.84 & 16.78 & 34.15 & 24.93 & 21.76 
            & 30.29 & 20.90 & 17.61 & 40.26 & 27.08 & 23.14 \\

            \monodgp~\cite{pu2024monodgp} & None 
            & 26.35 & 18.72 & 15.97 & 35.24 & 25.23 & 22.02 
            & 30.76 & 22.34 & 19.02 & 39.40 & 28.20 & 24.42 \\

            \monocop~\cite{zhang2025unleashing} & None 
            & 27.54 & 19.11 & 16.33 & 36.77 & 25.57 & 22.62 
            & 32.06 & 23.98 & 20.64 & 42.20 & 31.29 & 27.58 \\

            \textbf{\methodName (Ours)} & None
            & \textbf{29.52} & \textbf{20.29} & \textbf{17.93} 
            & \textbf{37.55} & \textbf{26.59} & \textbf{23.26} 
            & \textbf{33.61} & \textbf{24.40} & \textbf{20.80} 
            & \textbf{44.69} & \textbf{32.17} & \textbf{27.93} \\
            \midrule
            \bottomrule[1pt]
        \end{tabular}
    }
    \vspace{-1mm}
    \caption{\textbf{\kitti Leaderboard (\test) and \val results at \iouThreeD $\mathbf{\geq 0.7}$.} 
    \methodName achieves \sota performance across all metrics, demonstrating that our intrinsic aware design also improves standard \threeD benchmarks.}
    \label{exp:test-car}
    \vspace{-3mm}
\end{table*}



\subsection{Results on Multi-dataset Training}
\label{exp:multi-dataset}

Since \methodName is intrinsically aware, it can naturally integrate datasets captured with different focal lengths, enabling unified multi-dataset training. 
As shown in \cref{exp:combined-car}, existing detectors such as \monocop\ fail under heterogeneous intrinsics, dropping from \(23.64\% \rightarrow 17.26\%\) on \kitti\ and from \(7.39\% \rightarrow 6.21\%\) on \nuscenes. 
Applying virtual-depth (VD) normalization~\cite{brazil2023omni3d} alleviates but does not eliminate this degradation. 
In contrast, \methodName improves from \(24.40\%/8.12\%\) (single-dataset) to \(26.54\%/9.81\%\) (joint \kitti+\nuscenes) and further to \(28.91\%/11.48\%\) when scaled to three datasets (\kitti, \nuscenes, \waymo). 
These results confirm that our intrinsic-aware design bridges inter-dataset discrepancies and generalizes across diverse camera intrinsics. We provide more detailed results in Appendix\,\ref{appx:multi-data}

\vspace{-1mm}
\begin{table}[t]
\centering
        \centering
        \tabcolsep=0.1cm
        \resizebox{0.48\textwidth}{!}{
        \begin{tabular}{llcc}
            \toprule[1pt]
            \midrule
            Method & Trained on   &$AP_{3D}^{KIT}$ &  $AP_{3D}^{NU}$ \\
            \midrule
            \monocop~\cite{zhang2025unleashing} & KIT &  23.98 & \mathDash \\
            \monocop~\cite{zhang2025unleashing} & NU &  \mathDash & 7.39 \\
            \monocop~\cite{zhang2025unleashing} & KIT $+$ NU & 17.26 & 6.21  \\      
            \monocop~\cite{zhang2025unleashing} + VD~\cite{brazil2023omni3d} & KIT $+$ NU  & 23.15 &  7.42\\  
            \midrule
            
            \methodName & KIT & {24.40} & \mathDash \\
            \methodName & NU & \mathDash & {8.12}\\
             \methodName & KIT $+$ NU & {26.54} & {9.81}\\
            \methodName & KIT + NU + Way &\textbf{28.91} & \textbf{11.48} \\
            \midrule
            \bottomrule[1pt]
        \end{tabular}}
        \vspace{-2mm}
        \caption{\textbf{Multi dataset training results.}
        Our intrinsic aware design helps bridge inter dataset discrepancies and improves overall detection performance across \kitti and \nuscenes. 
        [Key: KIT = \kitti, NU = \nuscenes, Way = \waymo, VD = Virtual Depth]}
        \label{exp:combined-car}
        \vspace{-3mm}
\end{table}

\subsection{Results on Standard \threeD Benchmarks} 
\label{exp:standard}

\begin{table}[t]
    \centering
    \tabcolsep=0.3cm
    \resizebox{0.48\textwidth}{!}{
    \begin{tabular}{l|cccc}
        \toprule[1pt]
        \midrule
        \multirow{2}{*}{Method} & \multicolumn{2}{c}{\apThreeD} & \multicolumn{2}{c}{\apBev} \\
         & Easy & Mod. & Easy & Mod. \\
        \midrule
        \deviant~\cite{kumar2022deviant}  & 9.69 & 8.33 & 16.28 & 14.36 \\
        \monodetr~\cite{zhang2023monodetr}  & 9.53 & 8.19 & 16.39 & 14.41 \\
        \monodgp~\cite{pu2024monodgp}  & 10.04 &  8.78 & 16.55 & 14.53 \\ 
        \monocop~\cite{zhang2025unleashing}   & \underline{10.85} & \underline{9.71} & \underline{17.83} & \underline{15.86} \\
        \textbf{\methodName (Ours)}   & \textbf{12.33} & \textbf{10.74} & \textbf{19.56} & \textbf{17.33} \\
        \midrule
        \bottomrule[1pt]
    \end{tabular}}
    \caption{
        \textbf{\nuscenes \val Results.} \methodName achieves \sota performance on \threeD detection and \bev detection under IoU $\ge 0.7$. 
        [Key: \textbf{First}, \underline{Second}]
    }
    \label{sec:experiment-nuscenes-0} 
    \vspace{-3mm}
\end{table}

\noindent\textbf{\kitti Leaderboard (\test) Results.}
\cref{exp:test-car} presents the official \kitti test results for the Car at IoU $\geq 0.7$, with all numbers sourced from the \kitti leaderboard. \methodName achieves \sota performance in both \apThreeD and \apBev, surpassing all previous image-only methods.
Notably, under the Moderate level which is considered the primary criterion on \kitti, \methodName outperforms \monocop by $+1.18\%$ in \threeD detection and $+1.02\%$ in BEV detection. Remarkably, even when compared to models that utilize additional LiDAR or depth inputs (\eg, MonoTAKD and OPA-3D), \methodName still delivers superior results, highlighting the effectiveness of our intrinsic-aware design.
We also provide more detailed results on KITTI in Appendix~\ref{appx:kitti}.

\noindent\textbf{\kitti \val Results.}
Tab.~\ref{exp:test-car} shows \methodName achieves consistent \sota performance on the \kitti \val split. It surpasses the previous best method \monocop by $+0.42\%$ \apThreeD on the Moderate level and $+1.55\%$ on the Easy level, aligning with the trends on the official \kitti leaderboard. These results confirm the effectiveness of intrinsic-aware modeling.

\noindent\textbf{\nuscenes \val Results.} Tab.~\ref{sec:experiment-nuscenes-0} shows \methodName achieves \sota performance on the  \nuscenes \val dataset.  For instance, \methodName outperforms \monocop   by $+1.48\%$ on the \apThreeD Easy level.

\noindent\textbf{\waymo \val Results.}
\noindent\methodName also achieves \sota performance on the \waymo \val and  \nuscenes \val . 
Due to space limitations, detailed results of Waymo and \nuscenes are provided in the Appendix~\ref{appx:waymo} and \ref{appx:nuscenes} respectively.

\begin{table*}[t]
    \centering
    \resizebox{0.9\textwidth}{!}{
    \tabcolsep=0.3cm
    \begin{tabular}{c|c|c@{$\rightarrowRHD$}c|ccc|ccc}
    \toprule[1pt]
    \midrule
    \multirow{2}{*}{Changed} & \multirow{2}{*}{\makecell{Row \\ Index}}
    & \multicolumn{2}{c|}{\multirow{2}{*}{From $\rightarrowRHD$ To}} 
    & \multicolumn{3}{c|}{\apThreeD, \iou $\geq 0.7$} 
    & \multicolumn{3}{c}{\apThreeD, \iou $\geq 0.5$} \\
    & & \multicolumn{2}{c|}{} 
    & Easy & Mod. & Hard 
    & Easy & Mod. & Hard \\
    \midrule
    \multirow{2}{*}{Baseline} & 1 & \multicolumn{2}{c}{Single Focal} & 32.40 & \cellcolor{lightgray}23.64 & 20.31 & 71.30 & 54.70 & 48.66 \\
    & 2 &  \multicolumn{2}{c|}{Synthetic Images} & 29.77 & \cellcolor{lightgray}21.71 & 17.46 & 69.53 & 51.20 & 46.87 \\
    \midrule
    \multirow{2}{*}{Intrinsic Encoder} & 3 & Yes & No      		  & 29.80 & \cellcolor{lightgray}22.16 & 17.76 & 69.61 & 52.63 & 46.57 \\
    & 4 & Frozen & Trainable    & 29.76 & \cellcolor{lightgray}21.85 & 18.77 & 68.81 & 52.30 & 47.18 \\
    \midrule    
    Connector   & 5     & Yes & No        & 31.97 & \cellcolor{lightgray}22.85 & 19.40 & 69.54 & 52.04 & 48.25 \\
    
    Feature-Level Adaptation & 6  & Yes & No         & 32.48 & \cellcolor{lightgray}23.43 & 20.03 & 71.04 & 54.02 & 47.94 \\
    Query-Level Adaptation & 7    & Yes & No         & \textbf{34.02} & \cellcolor{lightgray}23.99 & 20.32 & 71.50 & 54.25 & 47.85 \\
    \midrule

    \rowcolor{lightgray} \textbf{\methodName (Ours)} & 8 & \multicolumn{2}{c|}{\mathDash} & 33.61 & \cellcolor{lightgray}\textbf{24.40} & \textbf{20.80} & \textbf{71.96} & \textbf{55.29} & \textbf{50.63} \\
    \midrule
    \bottomrule[1pt]
    \end{tabular}}
    \vspace{-1mm}
    \caption{\textbf{Ablation studies} on \kitti  validate the effectiveness of each module in enabling intrinsic-aware detection.}
    \label{exp:ablation}
    \vspace{-3mm}
\end{table*}

\begin{table}[t]
    \centering
    \tabcolsep=0.3cm
    \resizebox{0.46\textwidth}{!}{
        \begin{tabular}{lccc}
            \toprule[1pt]
            \midrule
            \multirow{2}{*}{Method} & \multirow{2}{*}{\apThreeD} &\multicolumn{2}{c}{Efficiency}  \\
            & & \#Param (M) & GFLOPs \\
            \midrule
            \monodetr~\cite{zhang2023monodetr} & 20.61 & 35.93 & 59.72 \\ 
            \monodgp~\cite{pu2024monodgp} & 22.34 & 38.90 & 68.99 \\
            \monocop~\cite{zhang2025unleashing} & 23.98 & 42.50 & 71.77 \\
            \textbf{\methodName (Ours)} & \textbf{24.40} & 42.63 & 71.77 \\   
            \midrule
            \bottomrule[1pt]
        \end{tabular}}
            \vspace{-2mm}
    \caption{\textbf{Efficiency comparison on the \kitti \val set.} 
    \methodName achieves the \sota performance while maintaining comparable model size and computational cost to prior works.}
    \vspace{-3mm}
    \label{exp:val-car-efficiency}
\end{table}

\subsection{Efficiency Analysis}
Beyond accuracy, as shown in \cref{exp:val-car-efficiency}, \methodName remains highly efficient. It introduces only a marginal increase of $+0.13$M parameters and identical GFLOPs compared to \monocop, yet yields consistent performance gains. This indicates that the improvement comes from a more effective design rather than increased model capacity.

\subsection{Ablation Study}

In this section, we conduct ablation studies to understand the effects of each component of \methodName on the \kitti \val set.  Unless otherwise specified, we adopt \apThreeD at $\text{IoU} \geq 0.7$ (Moderate) as the primary evaluation metric.
The results summarized in \cref{exp:ablation}, progressively reveal how (1) intrinsic simulation alone is insufficient, (2) semantic intrinsic encoding enables generalization, and (3) hierarchical adaptation and connector design further enhance the alignment between intrinsic knowledge and visual representations. 
We also provide additional ablations in Appendix~\ref{appx:more-ablations}.

\noindent\textbf{Intrinsic Simulation Module.}
We first examine whether the performance gain primarily comes from using the synthetic multi-intrinsic dataset.
As shown in \cref{exp:ablation} (Row 1 vs.~2), directly applying the synthetic data to the baseline \monocop leads to a $1.93\%$ performance drop, demonstrating that simply increasing data diversity without intrinsic-aware modeling is ineffective.
This highlights the necessity of explicitly encoding and adapting intrinsic information rather than relying on raw data augmentation alone.

\noindent\textbf{Intrinsic Encoder.}
We then evaluate the contribution of the Intrinsic Encoder.
As shown in \cref{exp:ablation} (Row 3), replacing the Intrinsic Encoder with a simple linear layer that directly encodes raw intrinsic values leads to a notable performance drop, confirming the importance of semantically meaningful intrinsic embeddings.
We further find that freezing the intrinsic embeddings during training is crucial for maintaining stable performance: without freezing, the accuracy drops from $24.40\%$ to $21.85\%$ (Row 4 in \cref{exp:ablation}).
This indicates that updating these embeddings distorts the semantic structure inherited from the pretrained LLM and CLIP encoders, while freezing them preserves the intrinsic knowledge necessary for effective generalization.

\noindent\textbf{Intrinsic Adaptation Module.}
Finally, we assess the effectiveness of the Intrinsic Adaptation Module.
Removing the Connector leads to a noticeable performance drop from $24.40\%$ to $22.85\%$ in \apThreeD, confirming its necessity for projecting intrinsic embeddings into a learnable feature space and aligning them with visual representations.
We further evaluate the two components of the hierarchical fusion design: removing the Feature-Level Adaptation (Row~6) results in the most significant degradation, while removing the Query-Level Adaptation (Row~7) causes a smaller yet consistent decline ($-0.41\%$).
These results demonstrate that both adaptation stages contribute to intrinsic-aware learning, with feature-level integration playing a more critical role in preserving geometric consistency.

\noindent\textbf{Visualizations.}
Appendix\,\ref{appx:vis} includes further visualizations of \methodName under \kitti, \nuscenes and \waymo.

\section{Conclusion}
\label{sec:conclusion}

We presented \methodName, a unified intrinsic-aware framework that converts numeric intrinsics into language-grounded representations capturing their perceptual and geometric implications, and integrates them hierarchically into the detection pipeline. This design enables detectors to interpret how intrinsic changes affect perception and adapt their features accordingly. 
Extensive experiments across multiple benchmarks show that \methodName generalizes well to unseen focal lengths, supports multi-dataset training, and achieves new \sota results. 
We believe modeling camera intrinsics as semantic representations offers a promising path toward geometry- and perception-aware \threeD vision systems that remain reliable across diverse real-world cameras.

{
    \small
    \bibliographystyle{ieeenat_fullname}
    \bibliography{main}
}

\clearpage
\setcounter{section}{0}
\setcounter{figure}{0}
\setcounter{table}{0}
\makeatletter 
\renewcommand{\thesection}{\Alph{section}}
\renewcommand{\theHsection}{\Alph{section}}
\renewcommand{\thefigure}{A\arabic{figure}}
\renewcommand{\theHfigure}{A\arabic{figure}}
\renewcommand{\thetable}{A\arabic{table}}
\renewcommand{\theHtable}{A\arabic{table}}
\makeatother
\renewcommand{\thetable}{A\arabic{table}}
\setcounter{equation}{0}
\renewcommand{\theequation}{A\arabic{equation}}

\section*{\Large{Appendix}}

In this appendix, we provide extended discussions and additional results that complement the main paper.
\cref{appx:syn-images} presents examples of the images generated by the Intrinsic Simulation Module.
\cref{appx:llm} shows examples of the intrinsic aware text descriptions generated by the Intrinsic Encoder.
\cref{appx:implementation} summarizes the hyperparameter settings and implementation details.
\cref{appx:multi-data} reports the results for multi dataset training.
\cref{appx:kitti} provides detailed results on the \kitti dataset.
\cref{appx:waymo} provides detailed results on the \waymo dataset.
\cref{appx:nuscenes} provides detailed results on the \nuscenes dataset.
\cref{appx:more-ablations} includes additional ablation studies.
\cref{appx:vis} provides further qualitative visualizations.
\cref{appx:limitation} discusses the limitations of \methodName.

\section{Generated Images}
\label{appx:syn-images}
\begin{figure*}[t]
    \centering
    \includegraphics[width=1\linewidth]{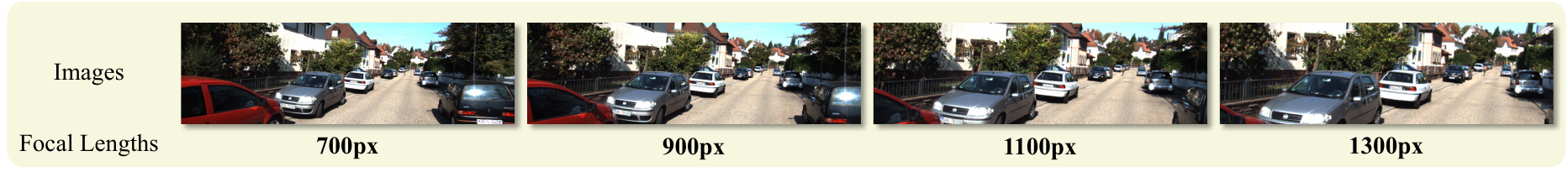}
    \caption{\textbf{Examples of synthetic images.} }
    \label{fig:images}
\end{figure*}

We first present the synthetic images produced by the Intrinsic Simulation Module. Given an original image and its associated ground truth focal length, the module renders a new view that corresponds to a target focal length. This is achieved by adjusting the field of view according to the target focal and then rescaling the transformed image back to the original resolution to maintain a consistent input size for the detector.
This process allows us to systematically vary the intrinsic parameters while preserving the scene content, enabling controlled studies of intrinsic sensitivity and robustness. Representative examples of the rendered images are shown in~\cref{fig:images}, where changes in effective perspective, object scale, and scene geometry become visually evident as the focal length varies.

\section{Generated Intrinsic Texts}
\label{appx:llm}

\begin{figure*}[t]
    \centering
    \includegraphics[width=1\linewidth]{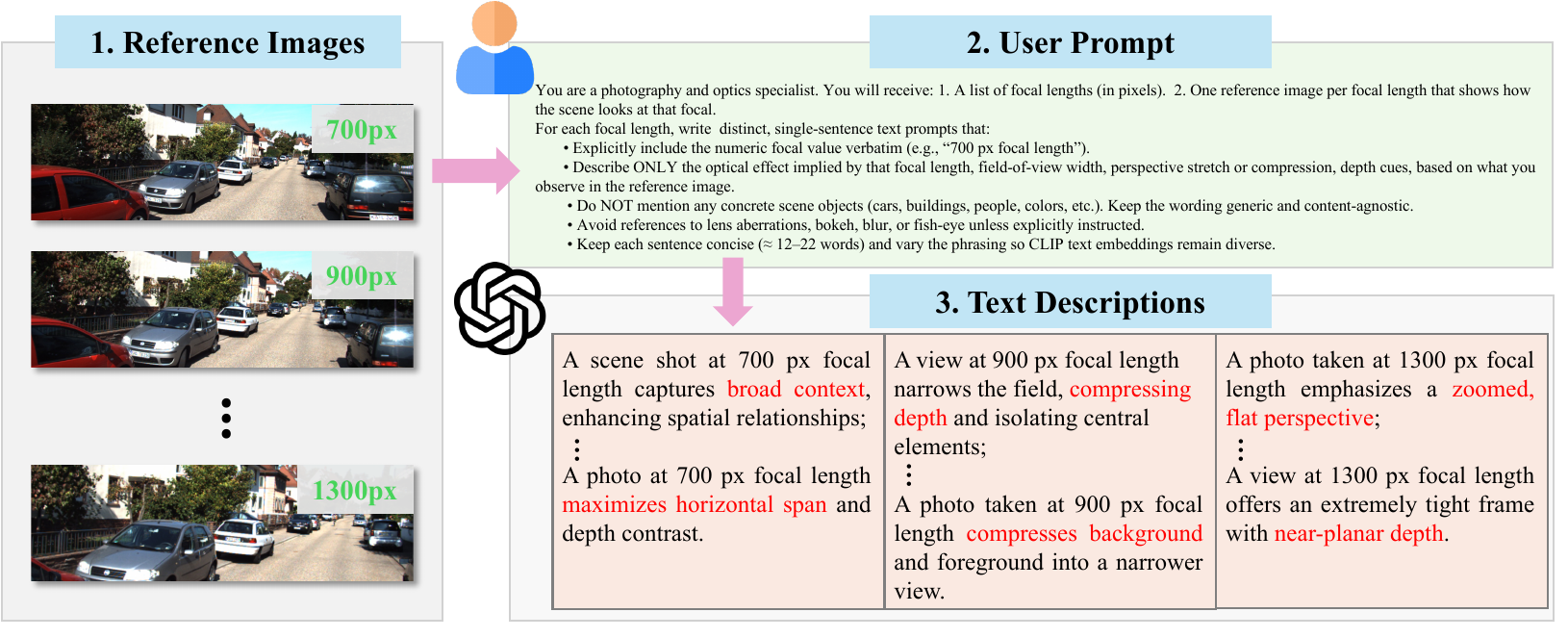}
    \caption{\textbf{Overiew of Generation of Intrinsic Texts.}  }
    \label{fig:text-llm}
\end{figure*}

In the Intrinsic Encoder, we take the simulated images as visual references and use them to guide the generation of intrinsic aware text descriptions. Specifically, for each pair of original and focal length transformed images, we provide both views to a large language model and ask it to articulate the visual effects introduced by the intrinsic change. The model describes how the modified focal length alters object scale, perceived depth, foreground background separation, and overall scene perspective.
These descriptions are phrased in natural language and capture the perceptual consequences of intrinsic variation rather than numeric changes alone. They serve as semantically rich prompts that enable the detector to associate visual cues with their underlying intrinsic causes.
We provide representative examples of the generated text prompts in~\cref{fig:text-llm}, which illustrate how the language model explains perspective changes, scale distortion, and depth variation induced by different focal configurations.

\begin{table}[t]
  \centering
  \begin{tabular}{l|l}
    \toprule
    \textbf{Item} & \textbf{Value} \\
    \midrule
    optimizer & AdamW \\
    learning rate & 2e-4 \\
    weight decay & 1e-4 \\
    number of feature scales & 4 \\
    hidden dim & 256 \\
    nheads & 8 \\
    number of encoder layers & 3 \\
    number of decoder layers & 3 \\
    encoder npoints & 4 \\
    decoder npoints & 4 \\
    number of group & 11 \\
    $\alpha$ in class loss & 0.25 \\
    class loss weight & 2 \\
    bbox loss weight & 5 \\
    GIoU loss weight & 2 \\
    3D center loss weight & 10 \\
    dim loss weight & 1 \\
    depth loss weight & 1 \\
    depth map loss weight & 1 \\
    scheduler & Step \\
    decay rate & 0.5 \\
    decay list & [85,125,165,205] \\
    dropout & 0.1 \\
    number of queries & 50 \\
    feedforward dim & 256 \\
    class cost weight & 2 \\
    bbox cost weight & 5 \\
    GIoU cost weight & 2 \\
    3D center cost weight & 10 \\
    \bottomrule
  \end{tabular}
  \caption{\textbf{Main hyperparameters of \methodName.}}
  \label{tab:hyper}
\end{table}

\section{Implementation details}
\label{appx:implementation}

\methodName is built upon the \monodgp~\cite{pu2024monodgp} and \monocop~\cite{zhang2025unleashing} frameworks. For each intrinsic configuration, we use ChatGPT 4o~\cite{openai2024chatgpt} to generate a collection of $24$ diverse and semantically informative prompts that describe the intrinsic properties in natural language. These descriptions are encoded using the CLIP ViT H/14~\cite{xue2023ulip} text encoder, which offers strong image text alignment and provides a stable foundation for intrinsic aware representation learning.
We summarize training hyperparameters in~\cref{tab:hyper}. We adopt the AdamW optimizer with a learning rate of $2 \times 10^{-4}$ and a weight decay of $10^{-4}$. The model is trained for $250$ epochs with a batch size of $16$.

\section{Multi-dataset Training Results}
\label{appx:multi-data}
Due to its intrinsic awareness, \methodName naturally supports multi-dataset training, a setting where images originate from cameras with vastly different intrinsic parameters such as focal lengths, sensor sizes, and principal point offsets. Conventional monocular 3D detectors struggle in this regime because they implicitly assume a fixed projection geometry. When trained on heterogeneous datasets, their learned depth–feature relationships become inconsistent, leading to degraded or unstable performance. In contrast, \methodName explicitly conditions the detection pipeline on intrinsic representations, allowing it to correctly interpret geometric variations across datasets and maintain consistent depth reasoning.

To examine this capability in depth, we evaluate several multi-dataset training strategies that vary category coverage and input resolutions. As shown in~\cref{exp:multi-dataset-results}, \methodName exhibits consistent improvements across all configurations, demonstrating its ability to absorb complementary information from different datasets while aligning their geometric discrepancies through intrinsic-aware modeling.
Specifically, when trained on all categories using both \kitti and \nuscenes, the \kitti \val performance improves from $24.40\%$ to $26.54\%$, while the \nuscenes performance increases from $7.09\%$ to $8.19\%$. This indicates that the model not only benefits from the additional visual diversity but also properly handles the large intrinsic gap between these two datasets. Furthermore, when \waymo is included in the joint training set, the \kitti performance further rises from $26.54\%$ to $28.91\%$. Notably, this gain persists despite the fact that \waymo has yet another distinct imaging pipeline and scene distribution, which typically destabilizes conventional monocular 3D detectors.

These results together reveal two key findings. First, multi-dataset training with \methodName yields a model that surpasses all individually trained models, suggesting that intrinsic-aware design enables effective consolidation of knowledge from different domains. Second, the unified model is not only more accurate but also more robust, highlighting that intrinsic conditioning allows \methodName to generalize across datasets, these experiments confirm that intrinsic-aware modeling provides a principled mechanism for exploiting visual diversity while preserving coherent depth representations, which is crucial for building scalable monocular 3D detectors in real-world multi-camera environments.

\begin{table*}[t]
\centering
        \centering
        \begin{tabular}{c|c|c|ccc|c|c|c}
            \toprule[1pt]
            \midrule
            \multirow{2}{*}{Method} & \multirow{2}{*}{\makecell{Training \\ Cates}} &\multirow{2}{*}{Resolution} & \multicolumn{3}{c|}{Training Datasets}   & \multirow{2}{*}{$AP_{3D}^{KIT}$} &  \multirow{2}{*}{$AP_{3D}^{NU}$} & \multirow{2}{*}{$AP_{3D}^{Way}$} \\
            & & & KITTI & nuScenes & Waymo & & &  \\
            \midrule
            \monodetr~\cite{zhang2023monodetr} & \multirow{2}{*}{Car} & \multirow{2}{*}{1280 $\times$ 384} & \cmark & & & 20.61\\ 
            \monodgp~\cite{pu2024monodgp} &  &  & \cmark & & & 22.49\\ 
            \midrule
            \monocop ~\cite{zhang2025unleashing}& \multirow{2}{*}{ALL} & \multirow{2}{*}{1280 $\times$ 384} & \cmark & & & 23.64\\ 
            \methodName &  &  & \cmark & & & 24.40\\ 
            \midrule
            \methodName & ALL & 896 $\times$ 512 &  & \cmark & & & 7.09  \\ 
            \methodName & ALL & 768 $\times$ 512 &  &  & \cmark & & & 8.94  \\ 
            \midrule
            \multirow{2}{*}{\methodName} & \multirow{2}{*}{ALL} & \multirow{2}{*}{1280 $\times$ 384} & \cmark & \cmark &  &  25.09 & 8.19 &  \\ 
            & & & \cmark & \cmark & \cmark & 26.28 & 10.45 & 7.77 \\ 
            \midrule
            \multirow{3}{*}{MonoIA} & ALL & \multirow{3}{*}{1280 $\times$ 512} & \cmark & \cmark &  &  26.54 & 9.81\\
             & ALL &  & \cmark & \cmark & \cmark & 28.91 & 11.48 & 10.19 \\
            & Car & & \cmark & \cmark & \cmark & 29.31 & 12.70 & 11.84 \\
            \midrule
            \bottomrule[1pt]
            
        \end{tabular}
        \caption{\textbf{Detailed results on Multi-Dataset Training.} [Key: KIT: KITTI, NU: nuScenes, Way: Waymo]}
        \label{exp:multi-dataset-results}
\vspace*{-2mm}
\end{table*}

\section{Detailed KITTI Results}
\label{appx:kitti}

\begin{table*}[h]
    \centering
    \begin{tabular}{l|c|c|ccc|ccc}
        \toprule[1pt]
        \midrule
        \multirow{2}{*}{Method} & \multirow{2}{*}{\makecell{Extra \\ Data}} & \multirow{2}{*}{Venue} 
        & \multicolumn{3}{c|}{\apThreeD (\uparrowRHDSmall)} 
        & \multicolumn{3}{c}{ \apBev (\uparrowRHDSmall)} \\
        & & & Easy & Mod. & Hard & Easy & Mod. & Hard \\
        \midrule
        \occupancymThreeD~\cite{peng2024learning} & LiDAR & CVPR~24 
        & 26.87 & 19.96 & 17.15 & 35.72 & 26.60 & 23.68 \\
        \opaThreeD~\cite{su2023opa} & Depth & ICRA~23 
        & 24.97 & 19.40 & 16.59 & 33.80 & 25.51 & 22.13 \\
        \midrule


        \monoflex~\cite{zhang2021objects} 
        & \multirow{11}{*}{\centering None} 
        & CVPR~21 
        & 23.64 & 17.51 & 14.83 & \mathDash & \mathDash & \mathDash \\

        \gupNet~\cite{lu2021geometry} 
        & 
        & CVPR~21 
        & 22.76 & 16.46 & 13.72 & 31.07 & 22.94 & 19.75 \\

        \deviant~\cite{kumar2022deviant} 
        & 
        & ECCV~22 
        & 24.63 & 16.54 & 14.52 & 32.60 & 23.04 & 19.99 \\

        \monocon~\cite{liu2022monocon} 
        & 
        & AAAI~22 
        & 26.33 & 19.01 & 15.98 & \mathDash & \mathDash & \mathDash \\

        \monouni~\cite{jia2023monouni} 
        & 
        & NeurIPS~23 
        & 24.51 & 17.18 & 14.01 & \mathDash & \mathDash & \mathDash \\

        \monodetr~\cite{zhang2023monodetr}  
        & 
        & ICCV~23 
        & 28.84 & 20.61 & 16.38 & 37.86 & 26.95 & 22.80 \\

        \monocd~\cite{yan2024monocd}  
        & 
        & CVPR~24 
        & 26.45 & 19.37 & 16.38 & 34.60 & 24.96 & 21.51 \\

        \fdThreeD~\cite{wu2024fd3d}  
        & 
        & AAAI~24 
        & 28.22 & 20.23 & 17.04 & 36.98 & 26.77 & 23.16 \\

        \monomae~\cite{jiang2024monomae}  
        & 
        & NeurIPS~24 
        & 30.29 & 20.90 & 17.61 & 40.26 & 27.08 & 23.14 \\

        \monodgp~\cite{pu2024monodgp}  
        & 
        & CVPR~25 
        & 30.76 & 22.34 & 19.02 & 39.40 & 28.20 & 24.42 \\

        \monocop~\cite{zhang2025unleashing}  
        & 
        & CVPR 26 
        & \underline{32.06} & \underline{23.98} & \underline{20.64} 
        & \underline{42.20} & \underline{31.29} & \underline{27.58} \\

        \textbf{\methodName (Ours)}  
        & 
        & CVPR 26 
        & \textbf{33.61} & \textbf{24.40} & \textbf{20.80} 
        & \textbf{44.69} & \textbf{32.17} & \textbf{27.93} \\

        \midrule
        \bottomrule[1pt]
    \end{tabular}
    \caption{\textbf{\kitti \val results at \iouThreeD $\ge 0.7$.} 
    \methodName achieves \sota~performance  across all metrics. [Key: \textbf{First}, Second]}
    \label{exp:val-car-detailed}
\end{table*}

While \cref{exp:test-car} in the main paper provides a simplified overview due to space constraints, we include a more comprehensive version here. \cref{exp:val-car-detailed} presents detailed comparisons of monocular \threeD object detection methods on the \kitti \val set under the challenging $\text{IoU} \ge 0.7$ setting. We report both \apThreeD and \apBev across Easy, Moderate, and Hard difficulty levels. \methodName achieves \sota performance across all metrics and difficulty levels, surpassing prior methods that either rely on external signals (\eg, depth or LiDAR) or use strong supervision across all categories.
Notably, our method maintains its superiority under both Car-only and All-category training settings. This highlights the robustness and generalizability of the proposed intrinsic-aware design, especially under data-scarce monocular settings. Additionally, while many competing methods incorporate external depth or LiDAR signals, \methodName achieves superior results with image-only input, demonstrating its efficiency and practicality. Interestingly, we observe that the most significant performance gain occurs at the Easy level. This aligns with the fact that enriching the dataset with diverse intrinsics increases the number of Easy objects, further validating the effectiveness of our Intrinsic Awareness in handling varying camera intrinsics.

\begin{table*}[t]
    \centering
    \begin{tabular}{c|c|l|cccc|cccc}
        \toprule[1pt]
        \midrule
        \multirow{2}{*}{\iouThreeD}  & \multirow{2}{*}{Difficulty} & \multirow{2}{*}{Method}  & \multicolumn{4}{c|}{\aphThreeD   \bracketPercentage (\uparrowRHDSmall)} & \multicolumn{4}{c}{\apThreeD{} \bracketPercentage (\uparrowRHDSmall)}\\ 
        
        &  & & All & 0-30 & 30-50 & 50-$\infty$ & All & 0-30 & 30-50 & 50-$\infty$  \\
        \midrule
        \multirow{14}{*}{0.7} & \multirow{7}{*}{Level 1} & \gupNet~\cite{lu2021geometry} in \cite{kumar2022deviant}  & 2.27 & 6.11 & 0.80 & 0.03 & 2.28 & 6.15 & 0.81 & 0.03 \\
        & & \deviant~\cite{kumar2022deviant} & 2.67 & 6.90 & 0.98 & 0.02 & 2.69 & 6.95 & 0.99 & 0.02   \\
        & & \monodetr~\cite{zhang2023monodetr} in \cite{zhang2025unleashing} & 2.10 & 5.94 & 0.73 & 0.12 & 2.11 & 5.99 & 0.73 & 0.12 \\
        & & \monodgp~\cite{pu2024monodgp} in \cite{zhang2025unleashing} & 2.39 & 6.62 & 0.84 & 0.12 & 2.41 & 6.67 & 0.84 & 0.12 \\
        & & \monocop~\cite{zhang2025unleashing} & \underline{2.70} & \underline{7.38} & \underline{1.06} & \textbf{0.16} & \underline{2.72} & \underline{7.44} & \underline{1.07} & \textbf{0.16} \\
         & & \textbf{\methodName (Ours)} & \textbf{3.05} & \textbf{8.43} & \textbf{1.11} & \underline{0.13} & \textbf{3.07} & \textbf{8.50} & \textbf{1.12} & \underline{0.13} \\
        \cmidrule(lr){2-11}
        & \multirow{7}{*}{Level 2} & \gupNet~\cite{lu2021geometry} in \cite{kumar2022deviant}   &   2.12 & 6.08 & 0.77 & 0.02 & 2.14 & 6.13 & 0.78 & 0.02 \\
        & & \deviant~\cite{kumar2022deviant}  & 2.50 & 6.87 & 0.94 & 0.02 & 2.52 & 6.93 & 0.95 & 0.02  \\
        & & \monodetr~\cite{zhang2023monodetr} in \cite{zhang2025unleashing} & 1.97 & 5.92 & 0.70 & 0.10 & 1.98 & 5.96 & 0.71 & 0.10 \\
        & & \monodgp~\cite{pu2024monodgp} in \cite{zhang2025unleashing} & 2.24 & 6.59 & 0.81 & 0.10 & 2.26 & 6.65 & 0.81 & 0.10  \\
        & & \monocop~\cite{zhang2025unleashing} & \underline{2.53} & \underline{7.35} & \underline{1.02} & \textbf{0.14} & \underline{2.55} & \underline{7.4}1 & \underline{1.03} & \textbf{0.14}  \\
        & & \textbf{\methodName (Ours)} & \textbf{2.86} & \textbf{8.40} & \textbf{1.07} & \underline{0.12} & \textbf{2.88} & \textbf{8.47} & \textbf{1.08} & \underline{0.12}  \\        
        \midrule
        \multirow{14}{*}{0.5} & \multirow{7}{*}{Level 1} & \gupNet~\cite{lu2021geometry} in \cite{kumar2022deviant} & 9.94 & 24.59 & 4.78 & 0.22 & 10.02 & 24.78 & 4.84 & 0.22 \\
        & &  \deviant~\cite{kumar2022deviant} & 10.89 & 26.64 & 5.08 & 0.18 & 10.98 & 26.85 & 5.13 & 0.18 \\
        & & \monodetr~\cite{zhang2023monodetr} in \cite{zhang2025unleashing} & 9.60 & 23.58 & 4.67 & 0.99 & 9.68 & 23.78 & 4.72 & 1.00 \\
        & & \monodgp~\cite{pu2024monodgp} in \cite{zhang2025unleashing} &  9.84 & 23.73 & 5.01 & 0.98 & 10.06 & 24.01 & 5.06 & 0.99 \\
        & & \monocop~\cite{zhang2025unleashing} &  \underline{11.65} & \underline{27.35} & \underline{5.97} & \textbf{1.46} & \underline{11.76} & \underline{27.59} & \underline{6.03} & \textbf{1.48} \\
        & & \textbf{\methodName (Ours)} &  \textbf{12.44} & \textbf{29.52} & \textbf{6.51} & \underline{1.18} & \textbf{12.54} & \textbf{29.75} & \textbf{6.57} & \underline{1.19} \\
        \cmidrule(lr){2-11}
        & \multirow{7}{*}{Level 2} & \gupNet~\cite{lu2021geometry} in \cite{kumar2022deviant} & 9.31 & 24.50 & 4.62 & 0.19 & 9.39 & 24.69 & 4.67 & 0.19 \\
        & & \deviant~\cite{kumar2022deviant} & 10.20 & 26.54 & 4.90 & 0.16 & 10.29 & 26.75 & 4.95 & 0.16 \\
        & & \monodetr~\cite{zhang2023monodetr} in \cite{zhang2025unleashing} & 9.00 & 23.49 & 4.51 & 0.86 & 9.08 & 23.70 & 4.55 & 0.87 \\
        & &  \monodgp~\cite{pu2024monodgp}  in \cite{zhang2025unleashing} & 9.32 & 23.65 & 4.84 & 0.85 & 9.43 & 23.92 & 4.88 & 0.86 \\
        & & \monocop~\cite{zhang2025unleashing}  & \underline{10.93} & \underline{27.25} & \underline{5.76} & \textbf{1.27} & \underline{11.03} & \underline{27.49} & \underline{5.82} & \textbf{1.29} \\
        & & \textbf{\methodName  (Ours)}  & \textbf{11.66} & \textbf{29.40} & \textbf{6.29} & \underline{1.03} & \textbf{11.76} & \textbf{29.64} & \textbf{6.34} & \underline{1.04} \\
        \midrule
        \bottomrule[1pt]
        \end{tabular}
    \caption{\textbf{\waymo \valOne Vehicle results.} \methodName  outperforms all methods on most metrics across both difficulty (Level $1$ and Level $2$) and \iou threshold ($0.5$ and $0.7$). [Key:  \textbf{First}, \underline{Second}]}
    \label{exp:mono3d_waymo}

\end{table*}

\section{Detailed Waymo Results}
\label{appx:waymo}
Table~\ref{exp:mono3d_waymo} presents a comprehensive comparison on the \waymo \valOne set under varying IoU thresholds ($0.5$ and $0.7$) and difficulty levels (Level $1$ and Level $2$). \methodName consistently outperforms all baselines across both \apThreeD and \aphThreeD, achieving top or second-best performance in every metric. Notably, our method demonstrates strong generalization across distances, 
particularly excelling in the mid-range $(0, 30\mathrm{m})$ and long-range $(30, 50\mathrm{m})$ detection, where prior approaches typically degrade. These results highlight the effectiveness of our intrinsic-aware design in enabling robust and accurate monocular \threeD object detection across challenging real-world scenarios.

\section{Detailed nuScenes Results}
\label{appx:nuscenes}
\cref{sec:experiment-nuscenes} summarizes our detection results on the \nuscenes \val split. Compared to existing methods, \methodName delivers strong improvements across both \threeD detection (\apThreeD) and BEV detection (\apBev) under multiple IoU thresholds. These gains are consistent across all difficulty levels, indicating that intrinsic-aware modeling provides more stable geometric reasoning in complex urban driving scenes.

Specifically, under the more challenging $\text{IoU} \ge 0.7$ setting, \methodName surpasses the strongest prior baseline, \monocop, by $+1.03\%$ on \apThreeD (Moderate) and $+1.47\%$ on \apBev (Moderate). Notably, \nuscenes includes diverse camera intrinsics and varying viewpoints, making high-IoU improvements particularly meaningful. Even under the relatively easier $\text{IoU} \ge 0.5$ condition, \methodName continues to outperform all baselines, achieving the highest scores across every category and metric. These results demonstrate that \methodName not only improves precise 3D localization but also enhances BEV spatial alignment, confirming its robustness in large-scale, multi-camera environments.

\begin{table*}[t]
    \centering
    \begin{tabular}{l|cccc|cccc}
        \toprule[1pt]
        \midrule
        \multirow{3}{*}{Method} &   \multicolumn{4}{c|}{\iouThreeD $\ge 0.7$} & \multicolumn{4}{c}{\iouThreeD $\ge 0.5$} \\
        & \multicolumn{2}{c|}{\apThreeD} & \multicolumn{2}{c|}{\apBev} & \multicolumn{2}{c|}{\apThreeD} & \multicolumn{2}{c}{\apBev} \\

         &   Easy & Mod. & Easy & Mod. & Easy & Mod. & Easy & Mod.\\
        \midrule
        \deviant~\cite{kumar2022deviant}  & 9.69 & 8.33 & 16.28 & 14.36 & 31.47 & 28.22 & 35.61 & 31.93\\
        \monodetr~\cite{zhang2023monodetr}  & 9.53 & 8.19 & 16.39 & 14.41 & 31.81 & 28.35 & 35.70 & 31.96\\
        \monodgp~\cite{pu2024monodgp}  & 10.04 &  8.78 & 16.55 & 14.53 &  29.56 & 26.17 & 32.67 & 29.44\\ 
        \monocop~\cite{zhang2025unleashing}   & \underline{10.85} & \underline{9.71} & \underline{17.83} & \underline{15.86}  & \underline{33.70} & \underline{29.91} & \underline{37.44} & \underline{34.01}\\
         \textbf{\methodName (Ours)}   & \textbf{12.33} & \textbf{10.74} & \textbf{19.56} & \textbf{17.33}  & \textbf{35.01} & \textbf{31.07} & \textbf{38.57} & \textbf{34.13}\\
        \midrule
        \bottomrule[1pt]
    \end{tabular}
    \caption{
        \textbf{\nuscenes \val Results.} \methodName  achieves \sota~performance on \threeD detection and \bev detection. [Key: \textbf{First}, \underline{Second}]}
        
    \label{sec:experiment-nuscenes} 
\end{table*}

\begin{figure*}[t]
    \centering
    \begin{subfigure}[t]{0.45\textwidth}  
        \includegraphics[width=\linewidth,keepaspectratio]{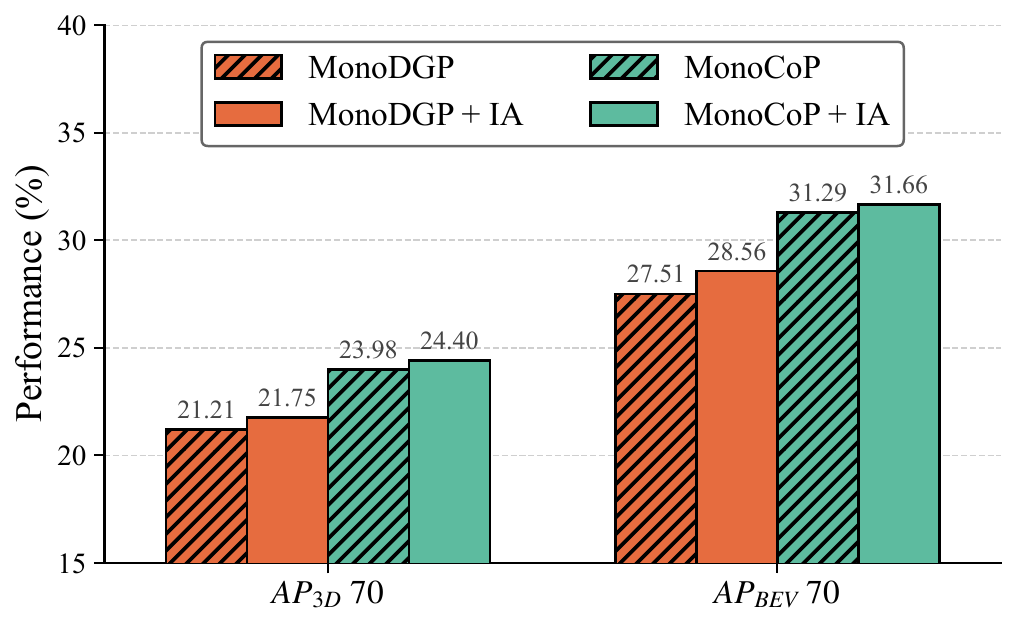}
        \caption{\apThreeDSeventy. performance.}
    \end{subfigure}
    \hfill
    \begin{subfigure}[t]{0.45\textwidth}  
        \includegraphics[width=\linewidth,keepaspectratio]{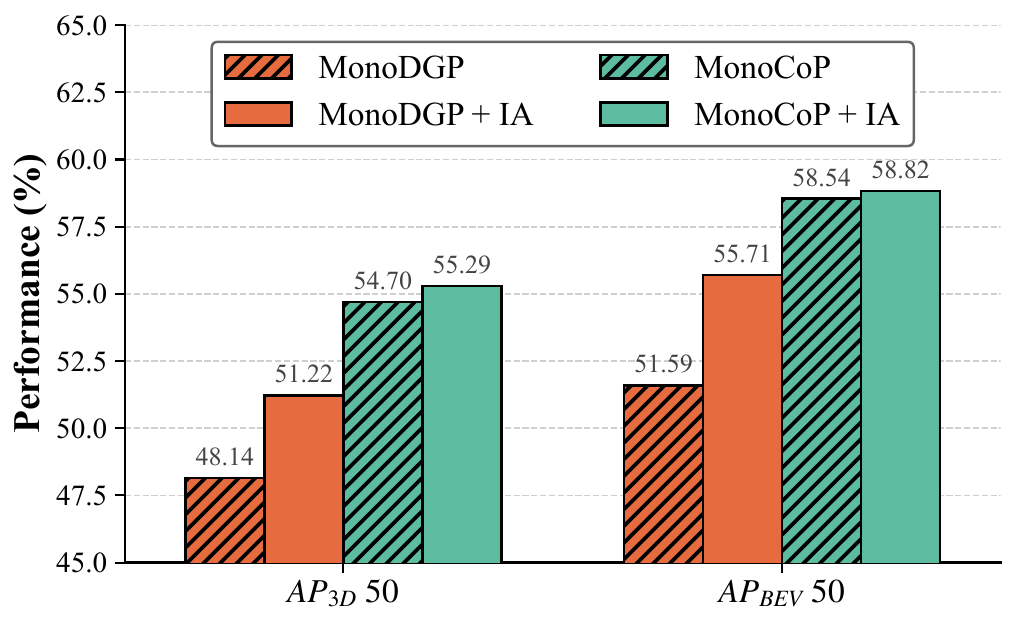}
        \caption{\apThreeDFifty performance.}
    \end{subfigure}
    \caption{Generalizability of Intrinsic Awareness (IA) on \textbf{different baseline methods}. }
    \label{appx:baseline}
\end{figure*}

\begin{figure*}[t]
    \centering
    \begin{subfigure}[t]{0.45\textwidth}  
        \includegraphics[width=\linewidth,keepaspectratio]{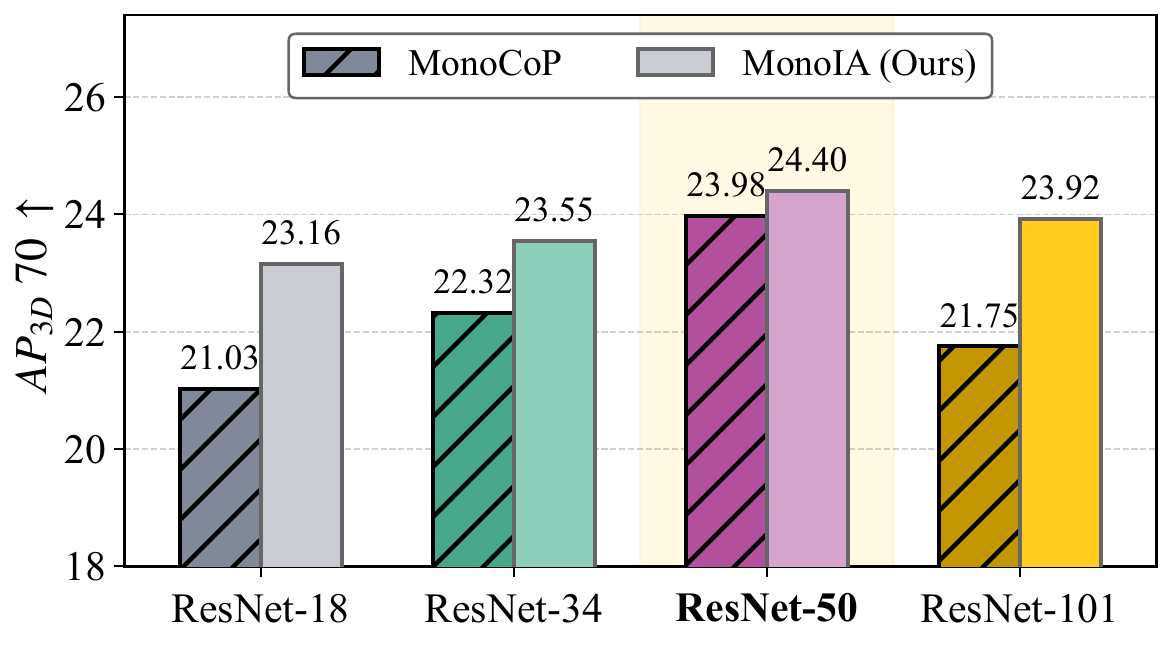}
        \caption{}
    \end{subfigure}
    \hfill
    \begin{subfigure}[t]{0.45\textwidth}  
        \includegraphics[width=\linewidth,keepaspectratio]{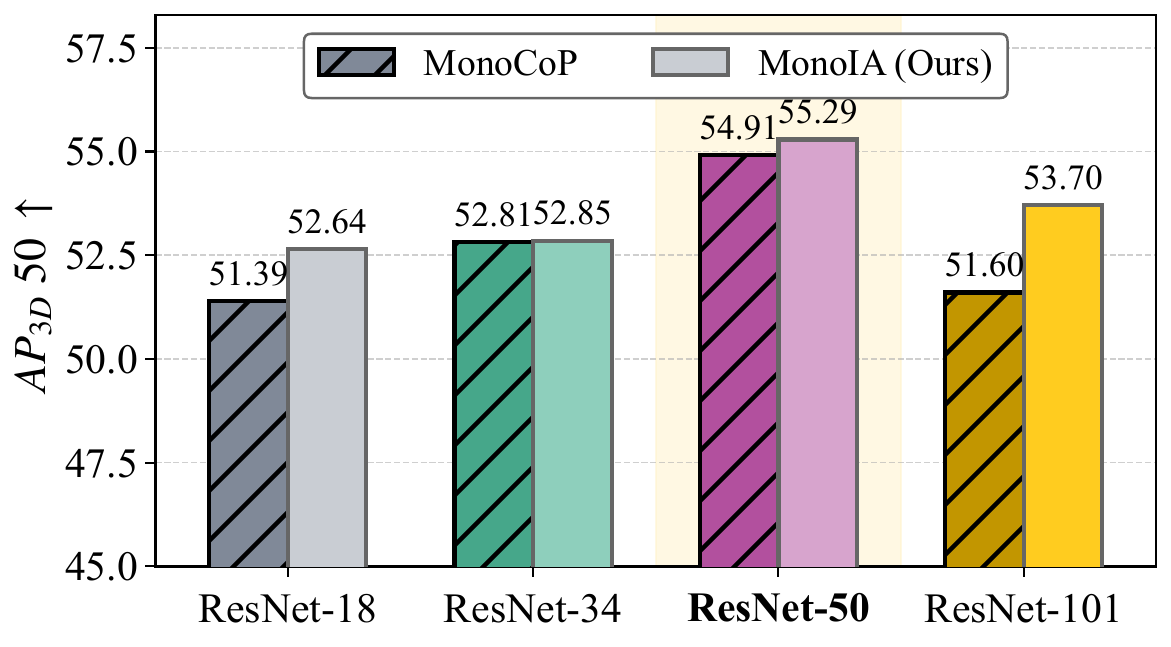}
        \caption{}
    \end{subfigure}
    \caption{Generalizability of Intrinsic Awareness (IA) on \textbf{different image backbones}. }
    \label{appx:backbones}
\end{figure*}
\section{More Ablations}
\label{appx:more-ablations}
We provide additional ablation studies to further assess the effectiveness of the proposed \methodName. 

\subsection{Support Multi baselines}
In the main paper, we primarily evaluate \methodName on the \monocop~\cite{zhang2025unleashing} framework. To further assess the generalizability and plug-and-play nature of our Intrinsic Awareness (IA) module, we additionally integrate it into \monodgp~\cite{pu2024monodgp}, a recent monocular \threeD detector accepted to CVPR 2025. As shown in~\cref{appx:baseline}, IA consistently improves both \apThreeD and \apBev across the two architectures. For example, \monodgp improves from $48.14\%$ to $51.22\%$ on \apThreeD{50} and from $51.59\%$ to $55.71\%$ on \apBev{50}. Similarly, \monocop improves from $54.70\%$ to $55.29\%$ on \apThreeD{50} and from $58.54\%$ to $58.82\%$ on \apBev{50} after adopting IA.

These consistent gains demonstrate two key properties of IA. First, IA enhances performance across detectors with very different designs. Second, the simultaneous improvements in both depth-sensitive metrics (\apThreeD) and BEV spatial metrics (\apBev) indicate that IA improves not only depth estimation but also the geometric coherence of the predicted 3D layout. Together, these results show that IA is a model-agnostic, easily pluggable module that can reliably strengthen a wide range of monocular \threeD detection frameworks.

\subsection{Support Multi-backbones.} 
Furthermore, we evaluate the robustness of \methodName across various image backbones, including ResNet-18, ResNet-34, ResNet-50, and ResNet-101. As shown in~\cref{appx:backbones}, \methodName consistently surpasses both \monocop and \monodgp across all backbones and difficulty levels. Notably, with a lightweight ResNet-18, \methodName achieves a significant gain of $+2.13\%$ in \apThreeD ($0.7$) Moderate over \monodgp. With deeper backbones like ResNet-50 and ResNet-101, \methodName maintains top performance, achieving $24.40\%$ and $23.92\%$ in Moderate settings, respectively. Interestingly, the slightly lower performance with ResNet-101 suggests that deeper networks do not always yield better results in \monoThreeD tasks. 
These consistent improvements demonstrate that our intrinsic-awareness design not only enhances performance but also generalizes effectively across architectures.

\subsection{Number of Intrinsic Texts}
We further study how the number of LLM generated intrinsic texts affects \methodName. In our design, these texts are encoded using the CLIP Text Encoder and then aggregated through average pooling to form the intrinsic text embedding. As shown in~\cref{tab:number-of-intrinsic-text}, increasing the number of intrinsic texts consistently improves performance, indicating that richer textual descriptions provide more stable intrinsic representations. We select $24$ texts as our default configuration, as it yields the best overall performance.
\begin{table}[t]
    \centering
    \begin{tabular}{c|ccc}
        \toprule[1pt]
        \midrule
         \multirow{2}{*}{\makecell{Number of\\Intrinsic Texts}} 
         & \multicolumn{3}{c}{\apThreeDSeventy (\%) (\uparrowRHDSmall)} \\
         & Easy & Mod & Hard \\
        \midrule
         0  & 29.80 & 22.16 & 17.76 \\
         1  & 32.53 & 23.69 & 20.09 \\
         12 & \textbf{33.96} & 24.17 & 20.55 \\
        \rowcolor{lightgray} 
         24 & 33.61 & \textbf{24.40} & 20.80 \\
         36 & 33.64 & 23.41 & \textbf{20.83} \\ 
        \midrule
        \bottomrule[1pt]
    \end{tabular}
    \caption{\textbf{Impact of Number of Intrinsic Texts.}}
    \label{tab:number-of-intrinsic-text}
\end{table}

\begin{table}[h]
    \centering
    \begin{tabular}{c|ccc}
        \toprule[1pt]
        \midrule
         \multirow{2}{*}{\makecell{Learned\\Tokens}} 
         & \multicolumn{3}{c}{\apThreeDSeventy (\%) (\uparrowRHDSmall)} \\
         & Easy & Mod & Hard \\
        \midrule
         {[700]}                    & 31.78 & 23.26 & 19.82 \\
         {[700, 900]}               & 32.32 & 23.54 & 20.14 \\
         {[700, 900, 1100]}         & 33.68 & 24.13 & 20.72 \\
        \rowcolor{lightgray}
         {[700, 900, 1100, 1300]}   & \textbf{33.61} & \textbf{24.40} & \textbf{20.80} \\
         {[700, 900, 1100, 1300, 1500]} & 32.36 & 23.41 & 19.93 \\
        \midrule
        \bottomrule[1pt]
    \end{tabular}
    \caption{\textbf{Impact of Number of Intrinsic Embeddings.}}
    \label{appx:impact_token}
\end{table}

\subsection{Number of Intrinsic Tokens}
We further evaluate the impact of the learned intrinsic tokens by training \methodName under different settings of intrinsics. As shown in~\cref{appx:impact_token}, increasing the number of intrinsic tokens initially improves performance. The model achieves peak performance when using four tokens, corresponding to a balanced representation of intrinsic variations. However, introducing more tokens beyond this point leads to a performance drop, due to the over-fragmentation of the intrinsic space.

\section{Visualization}
\label{appx:vis}
\begin{figure*}[h]
    \centering
    \includegraphics[width=1\linewidth]{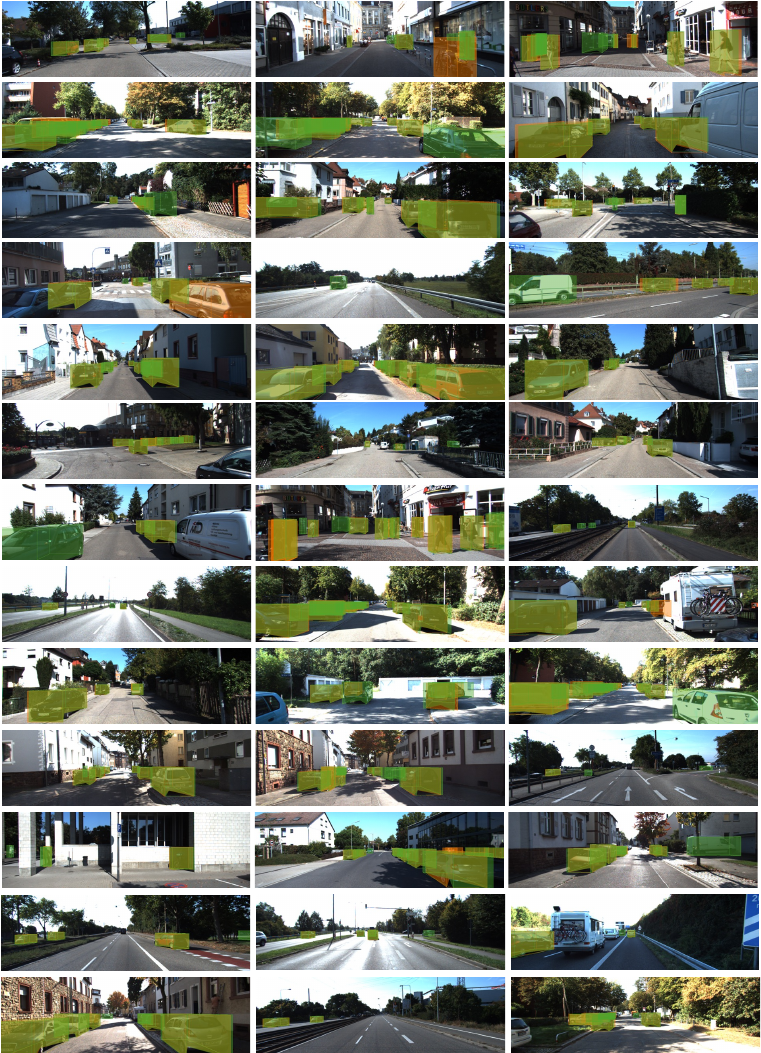}
    \caption{\textbf{Qualitative results on \kitti.} [Key: \textcolor{green}{\methodName}, \textcolor{orange}{\monocop}]  }
    \label{fig:vs-kitti-more}
\end{figure*}

\begin{figure*}[h]
    \centering
    \includegraphics[width=1\linewidth]{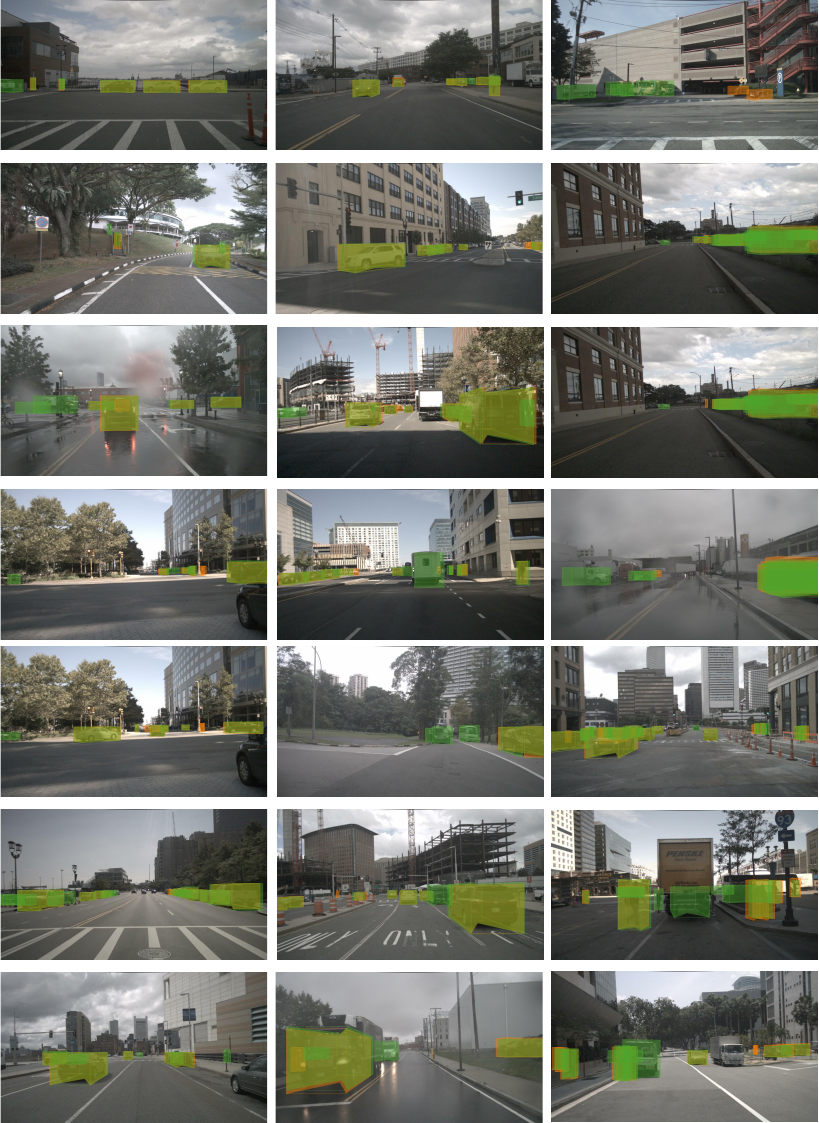}
    \caption{\textbf{Qualitative results on \nuscenes.}[Key: \textcolor{green}{\methodName}, \textcolor{orange}{\monocop}]   }
    \label{fig:nuscnee}
\end{figure*}

\begin{figure*}[t]
    \centering
    \includegraphics[width=1\linewidth]{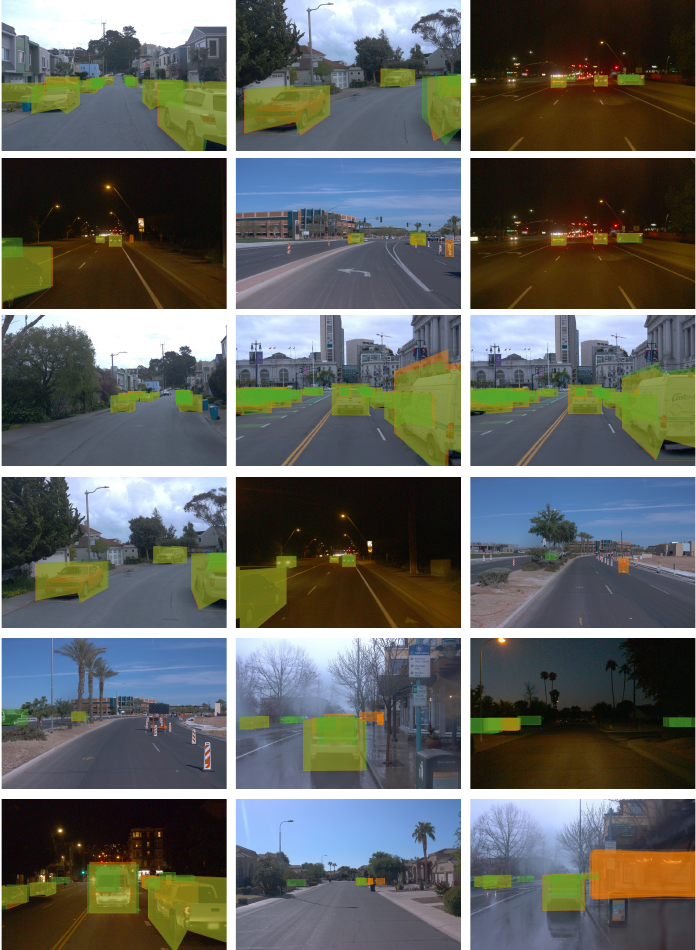}
    \caption{\textbf{Qualitative results on \waymo.}  [Key: \textcolor{green}{\methodName}, \textcolor{orange}{\monocop}] }
    \label{fig:waymo}
\end{figure*}

In this section, we present visualizations of detection results on \kitti (\cref{fig:vs-kitti-more}), \nuscenes (\cref{fig:nuscnee}), and \waymo (\cref{fig:waymo}). Predictions by the baseline method \monocop are highlighted in orange, while those by our \methodName are highlighted in green.

\section{Why mostly focusing on focal length variation.}
 We mainly focus on focal length for the following three reasons:

\noindent \textit{1) Geometric motivation.} 
Focal length is the dominant intrinsic component in Mono3D, as it directly controls the depth–scale mapping in monocular projection, while principal point shifts mainly induce image-plane translations. Such translation effects can be largely compensated by modern CNN- or Transformer-based detectors and therefore have a much smaller impact on Mono3D performance.

\noindent \textit{2) Experimental evidence.}
To support our choice, we separately evaluate principal point and focal length shifts on KITTI.  a 200-pixel principal point shift causes only marginal performance degradation, while an equivalent focal length shift leads to a much larger drop, indicating the dominant impact of focal length in \monoThreeD.

\noindent \textit{3) Cross-dataset practice.} 
Across common Mono3D benchmarks, the principal point is typically close to the image center and varies little, whereas intrinsic differences across datasets are mainly reflected in focal length. 
Therefore, focal length dominates intrinsic variation in practice.

\section{\methodName Limitation and Future Work}
\label{appx:limitation}

\methodName achieves intrinsic awareness by learning dedicated intrinsic embeddings. While it demonstrates strong generalization to unseen intrinsics, it is not an intrinsic-invariant network. Future work could explore intrinsic-invariant architectures that can naturally handle diverse camera settings without relying on explicit embedding learning.
Moreover, recent advances in multimodal learning and recognition systems have shown strong capabilities in visual reasoning and multimodal fusion~\cite{zhu2026fusionagent,chen2023atm,zhu2025quality,zhu2026can,su2026localscore,guo2026holistic,chen2025unlearning,chen2025safety}. However, these models are not designed for 3D perception tasks. Bridging vision-language models with 3D object detection, especially in terms of geometry-aware reasoning and spatial understanding, remains an important direction for future research.

\end{document}